\documentclass{article} 
\usepackage{iclr2021_conference,times}

\usepackage{amsmath,amsfonts,bm}

\def\eqref#1{equation~\ref{#1}}

\def\1{\bm{1}}

\DeclareMathAlphabet{\mathsfit}{\encodingdefault}{\sfdefault}{m}{sl}
\SetMathAlphabet{\mathsfit}{bold}{\encodingdefault}{\sfdefault}{bx}{n}

\usepackage[colorlinks = true,
            linkcolor = blue,
            urlcolor  = blue,
            citecolor = blue,
            anchorcolor = blue]{hyperref}
\usepackage{url}

\usepackage{graphicx}
\usepackage{subcaption}
\usepackage{enumitem}
\usepackage[most]{tcolorbox}
\usepackage{xcolor}
\usepackage{colortbl}
\usepackage{makecell}
\usepackage{wrapfig}
\usepackage{booktabs}
\usepackage{overpic}

\usepackage{algorithm}
\usepackage{algpseudocode}
\newcommand{\STATEnonum}{\item[]}
\usepackage{mathrsfs}  
\usepackage{arydshln}
\usepackage{comment}

\title{Generating Synthetic Clinical Data that\\
Capture Class Imbalanced Distributions with Generative Adversarial Networks:\\
Example using\\
Antiretroviral Therapy for HIV}

\author{Nicholas I-Hsien Kuo$^{1}$,\\
\textbf{Federico Garcia}$^{2}$, \textbf{Anders S\"onnerborg}$^{3}$, \textbf{Maurizio Zazzi}$^{4}$, \textbf{Michael B\"ohm}$^{5}$, \textbf{Rolf Kaiser}$^{5}$,\\
\textbf{Mark Polizzotto}$^{6}$, \textbf{Louisa Jorm}$^{1}$, \textbf{Sebastiano Barbieri}$^{1}$\\
$^{1}$Centre for Big Data Research in Health, the University of New South Wales, Sydney, Australia\\
$^{2}$Hospital Universitario San Cecilio, Granada, Spain\\
$^{3}$Hospital Karolinska Institutet, Stockholm, Sweden\\
$^{4}$Universit{\`a} degli Studi di Siena, Siena, Italy\\
$^{5}$Uniklinik K{\"o}ln, Universit{\"a}t zu K{\"o}ln, Cologne, Germany\\
$^{6}$Australian National University, Canberra, Australia\\
\textcolor{white}{*}\\
Corresponding author: Nicholas I-Hsien Kuo (\texttt{n.kuo@unsw.edu.au})\\
}

\iclrfinalcopy 
\begin{document}

\maketitle
\begin{abstract}
Clinical data usually cannot be freely distributed due to their highly confidential nature and this hampers the development of machine learning in the healthcare domain. One way to mitigate this problem is by generating realistic synthetic datasets using generative adversarial networks (GANs). However, GANs are known to suffer from mode collapse thus creating outputs of low diversity. This lowers the quality of the synthetic healthcare data, and may cause it to omit patients of minority demographics or neglect less common clinical practices. In this paper, we extend the classic GAN setup with an additional variational autoencoder (VAE) and include an external memory to replay latent features observed from the real samples to the GAN generator. Using \textit{antiretroviral therapy for human immunodeficiency virus} (ART for HIV) as a case study, we show that our extended setup overcomes mode collapse and generates a synthetic dataset that accurately describes severely imbalanced class distributions commonly found in real-world clinical variables. In addition, we demonstrate that our synthetic dataset is associated with a very low patient disclosure risk, and that it retains a high level of utility from the ground truth dataset to support the development of downstream machine learning algorithms.\\

\end{abstract}

\section*{Ethics Statement \& Reproducibility}
This study was approved by the University of New South Wales' human research ethics committee (application HC210661). We based our synthetic HIV dataset on EuResist~\citep{zazzi2012predicting}. For people in the EuResist integrated database, all data providers obtained informed consent for the execution of retrospective studies and inclusion in merged cohorts \citep{prosperi2010antiretroviral}.

The EuResist Integrated DataBase (EIDB) can be accessed for scientific studies once a proposal for analysis has been approved by EuResist's Scientific Board (see: \url{http://engine.euresist.org/database/}). To facilitate future research, our code will be made available through \url{https://github.com/Nic5472K/ScientificData2021_HealthGym}. Our synthetic dataset is freely accessible through \url{https://healthgym.ai/}.

\section{Introduction}\label{Sec:Introduction}

\begin{table*}[h!]
    \centering
    \begin{tabular}{|l|l|}
        \hline
        \textbf{Problem:} & 
        Synthetic clinical data can be used in place of highly confidential\\
        & real data. However, their quality greatly impacts\\
        & their utility to support the development of downstream models.\\
        \hline
        \textbf{What is Already Known:} & \textit{Generative adversarial networks} (GANs) can generate realistic\\
        & synthetic data but they are notorious for \textit{mode collapse} -- thus\\
        & drastically decreasing the cohort diversity of generated data.
        \\
        \hline
        \textbf{What this Paper Adds:} &
        We propose a technique to mitigate mode collapse in generating\\
        & clinical time series data. Of note, our synthetic data quality\\
        & were particularly marked for data related to minority groups.\\
        \hline
    \end{tabular}
\end{table*}

Advances in machine learning research for healthcare are hampered by an insufficient amount of openly available datasets. Clinical datasets are usually not readily accessible due to their confidential nature, and there is potential for patient harm if members of the cohort were to be successfully re-identified by an adversary \citep{el2020evaluating}. One way to manage this risk is to generate realistic synthetic datasets that are sufficiently similar to their real counterparts \citep{goncalves2020generation}. There is longstanding research in the field of synthetic data generation \citep{fienberg1998disclosure, walonoski2018synthea}; and more recently, \citet{li2021generating} and \citet{kuo2022health} showed promising results on generating mixed-type clinical time series data based on \textit{generative adversarial networks} (GANs) \citep{goodfellow2014generative, arjovsky2017wasserstein, gulrajani2017improved}.

Unlike most generative models \citep{kingma2014auto, sohl2015deep, van2016pixel}, GANs \citep{goodfellow2014generative} do not explicitly compute the data likelihood. Instead, they employ two sub-networks to solve a minimax game -- a generator to synthesise data from a random latent prior, and a discriminator to distinguish real data from the synthetic data. When an optimal discriminator is employed and the divergence can be minimised between the real and the generated data distributions, the generator learns to model complex probability distributions. 

While GANs have enjoyed a large amount of success in image generation \citep{yu2018generative, karras2019style} and their applications can also be found in both the natural language processing of text \citep{xu2018diversity} and of speech \citep{pascual2017segan}, they are not extensively studied in the medical domain. In fact, \citet{goncalves2020generation} found that the promising MedGAN approach \citep{choi2017generating, camino2018generating} performed unfavourably on the log-cluster metric \citep{woo2009global} (see Section \ref{Sec:Metrics}) indicating that 
their synthetic dataset was not sufficiently diversified\footnote{Refer to Table 7 and Figure 7 on page 16 of \citet{goncalves2020generation}. MedGAN failed to generate a synthetic dataset that faithfully represented multivariate categorical cancer data.}.

A synthetic dataset that is highly representative of its real counterpart could be used to develop \textit{reinforcement learning} (RL)~\citep{sutton2018reinforcement} algorithms. An RL agent could manage patient conditions from learning in an evolving clinical environment to construct a behaviour policy that optimises the duration, dosage, and type of treatment over time. However, learning from under-diversified synthetic datasets can lead to heavy biases in models and undermine fairness in patient care~\citep{bhanot2021problem}.

Generating realistic synthetic medical datasets is not a trivial task; and we attribute the difficulty to 
\begin{itemize}[noitemsep,topsep=0pt]
    \item the highly sparse and strongly correlated nature of variables in real world clinical datasets (see Section \ref{Sec:SparseNCorrelated}), and
    
    \item the tendency of mode collapse to occur during GAN training (see Section \ref{Sec:ModeCollapse}).
\end{itemize}

\begin{figure*}[ht!]
    \centering

    \includegraphics[width=0.9\linewidth]{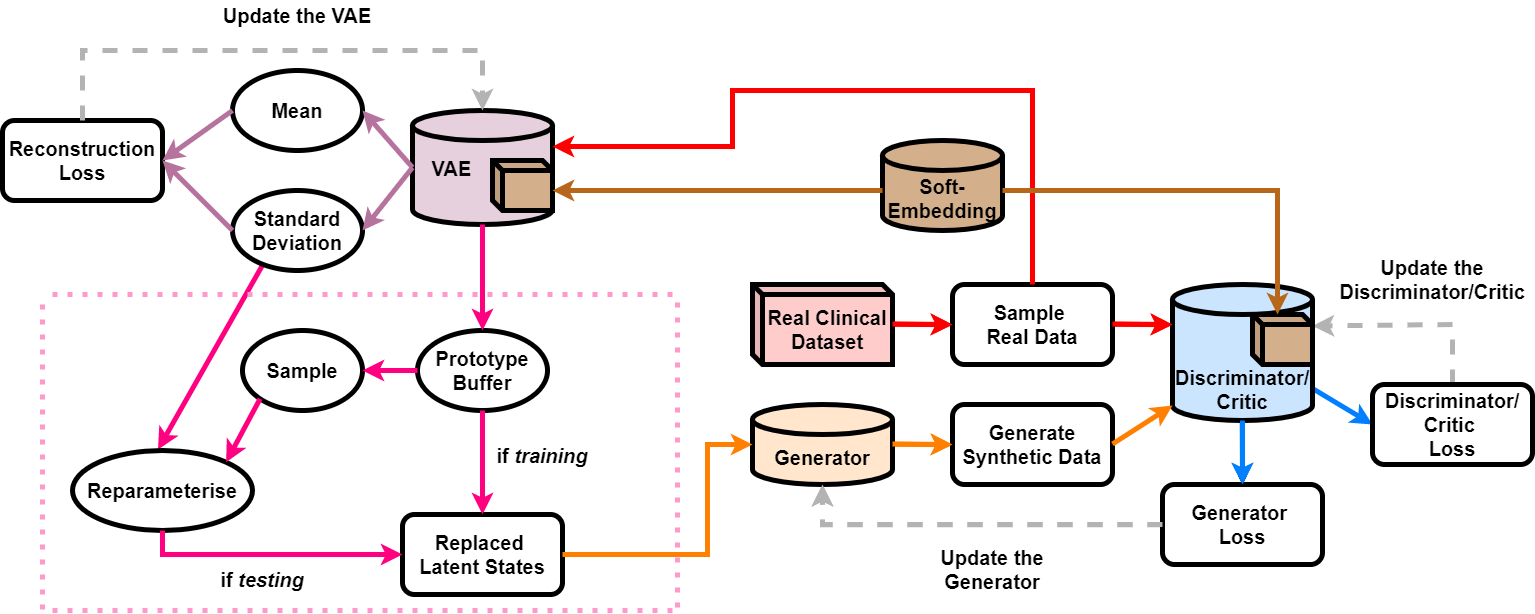}

    \caption{Extending GAN with a buffer replaying observed real data features.}\label{Fig:OurDesign}
\end{figure*}

We demonstrate that this algorithmic complication can be addressed by preconditioning the generator input. As illustrated in Figure \ref{Fig:OurDesign}, we extend the classic GAN setup with a \textit{variational autoencoder} (VAE) \citep{kingma2014auto} and a buffer to externally store features encoded from the real data by the VAE encoder. Whereas the classic setup uses random values \citep{goodfellow2014generative}, we sample and reparameterise features stored in the buffer as alternative generator inputs. For the rest of this paper, we will denote:\\
\hspace*{5mm} $\mathfrak{D}_\text{real}$ as the ground truth dataset;\\
\hspace*{5mm} $\mathfrak{D}_\text{null}$ as the null synthetic dataset generated using the classic GAN setup; and\\
\hspace*{5mm} $\mathfrak{D}_\text{alt}$ as the alternative synthetic dataset generated using our modified GAN setup.

In this paper, we use data related to \textit{antiretroviral therapy for human immunodeficiency virus} (ART for HIV) as a case study. Optimising ART for HIV is non-trivial because a person with HIV usually requires changes in treatment regimens to avoid the development of drug-resistant viral strains (see \cite{bennett2008world} and Chapter 4 of a guideline published by the \cite{world2016consolidated}). We observed that GAN generators with the classic random inputs synthesised \textit{overly structured} outputs and cannot capture the diversity in the combination of medications in ART for HIV regimen (see Section \ref{Sec:OverlyStructured}). We found that not only did our extended GAN setup (see Section \ref{Sec:OurShit}) increase the convergence rate, but it also captured severe class imbalanced combinations commonly found in real world clinical variables (see Section \ref{Sec:ResultsHIV}). 

This significantly boosted the utility of the synthetic dataset (see Section \ref{Sec:Uti02Z}). We found that RL agents trained on $\mathfrak{D}_\text{real}$ and $\mathfrak{D}_\text{alt}$ would suggest similar clinical actions to manage health states in people with HIV. However, this was not achieved for an RL agent trained on $\mathfrak{D}_\text{null}$. Our extended setup hence yielded a more reliable synthetic dataset which can be used to prototype RL algorithms in place of the highly confidential real data.

\section{Background}
This section discusses the ground truth ART for HIV dataset, the practical problems in training GANs, and compares the recent work of \citet{li2021generating} and \citet{kuo2022health} on successfully synthesising mixed type clinical datasets with GANs.

\subsection{The Ground Truth ART for HIV Dataset}\label{Sec:GroundTruth}

\begin{table}[ht!]
    \centering
    \begin{tabular}{|l||l|l|l|}
        \hline
        \textbf{Variable Name} & 
        \textbf{Data Type} & \textbf{Unit} &
        \textbf{Valid Categorical Options}\\
        
        \hline
        \hline
        Viral Load (VL) & 
        \cellcolor{blue!10}numeric & copies/mL & - -\\
        
        \hline
        Absolute Count for CD4 (CD4) & 
        \cellcolor{blue!10}numeric & cells/$\mu$L & - -\\

        \hline
        Relative Count for CD4 (Rel CD4) & 
        \cellcolor{blue!10}numeric & cells/$\mu$L & - -\\
        
        \hline
        \hline
        Gender &  
        \cellcolor{orange!10}binary & - - & \small{Male; \quad Female}\\
        
        \hline
        Ethnicity (Ethnic) &  
        \cellcolor{red!10}categorical & - - & \small{Asian; \quad African;}\\
        & & & \small{Caucasian; \quad Other}\\
        
        \hline
        \hline
        Base Drug Combination &  
        \cellcolor{red!10}categorical & - - & \small{FTC + TDF; \quad 3TC + ABC;}\\
        (Base Drug Combo)
        & & & 
        \small{FTC + TAF;}\\
        & & & 
        \small{DRV + FTC + TDF;}\\
        & & & 
        \small{FTC + RTVB + TDF; \quad Other}\\
        
        \hline
        Complementary INI &  
        \cellcolor{red!10}categorical & - - & \small{DTG; \quad RAL}\\
        (Comp. INI)
        & & & 
        \small{EVG; \quad Not Applied}\\
        
        \hline
        Complementary NNRTI &  
        \cellcolor{red!10}categorical & - - & \small{NVP; \quad EFV}\\
        (Comp. NNRTI) & & & 
        \small{RPV; \quad Not Applied}\\
        
        \hline
        Extra PI &  
        \cellcolor{red!10}categorical & - - & \small{DRV; \quad RTVB;}\\
        & & & 
        \small{LPV; \quad RTV;}\\
        & & & 
        \small{ATV; \quad Not Applied}\\
        
        \hline
        Extra pk Enhancer (Extra pk-En) &  
        \cellcolor{orange!10}binary & - - & 
        \small{False; \quad True}\\
        
        \hline
        \hline
        VL Measured (VL (M)) &  
        \cellcolor{orange!10}binary & - - & 
        \small{False; \quad True}\\
        
        \hline
        CD4 (M) &  
        \cellcolor{orange!10}binary & - - & 
        \small{False; \quad True}\\
        
        \hline
        Drug Recorded (Drug (M)) &  
        \cellcolor{orange!10}binary & - - & 
        \small{False; \quad True}\\

        \hline
    \end{tabular}
    
    \caption{\label{Tab:VarsOfHIV} The Variables of the ART in HIV Datasets
}
\end{table}

We selected a cohort of 8,916 people (with 332,800 records) from the EuResist database \citep{zazzi2012predicting} using published inclusion/exclusion criteria \citep{kuo2022health}. There are 3 numeric, 5 binary, and 5 categorical variables, as listed in Table \ref{Tab:VarsOfHIV}. The numeric variables --  VL, CD4, and Rel CD4 -- are indicative of the patient's health status. The HIV treatment regimens are deconstructed as Base Drug Combo, Complimentary (Comp.) INI, Comp. NNRTI, Extra PI, and Extra pk-En. The medication classes are: nucleoside reverse transcriptase inhibitors (NRTIs), nucleotide reverse transcriptase inhibitors (NtRTIs), non-nucleotide reverse transcriptase inhibitors (NNRTIs), integrase inhibitor (INI), and protease inhibitors (PIs). The base drug combo mainly comprises NRTIs + NtRTIs; see \citet{tang2012hivdb} for more details on the individual medications. There are 50 medication combinations in the real dataset spanning 21 medications.

The original EuResist database also contains a considerable proportion of missing data. Such information can be highly informative, indicating \textit{e.g.,} the need for specific laboratory tests; and hence we include the variables with suffix (M) to indicate \textbf{if measurements are taken}. Measurements are taken (\textit{i.e.,} entry with True) 24.27\% for VL (M), 22.21\% for CD4 (M), and 85.13\% for Drug (M) in the real dataset $\mathfrak{D}_\text{real}$. 

Data in EuResist are collected irregularly hence we summarised it across calendar months (taking the last reported value for each variable of that month). There are often long gaps (over 6 months) in the original records; and hence we split such records into multiple shorter sub-records. We truncate the sub-records' lengths to the closest multiple of ten; as a result, the shortest record has 10 months of data and the longest has 100 months. Time series data including medication usage are valuable for developing algorithms to optimise illness management. A similar dataset was used in \citet{parbhoo2017combining} to develop a hybrid approach of kernel-based regression and reinforcement learning for therapy selection. In the current paper, the generated synthetic data span 60 months of therapy for every patient, but the proposed setup can also be used to generate time-series of variable lengths.

\subsection{The Highly Sparse and Strongly Correlated Nature of Clinical Datasets}\label{Sec:SparseNCorrelated}

\begin{table}[ht!]
  \centering
  \begin{tabular}{l|lll}
    \toprule
    \texttt{Option} & Backbone & \colorbox{yellow!25}{Comp. NNRTI} & \colorbox{cyan!25}{Comp. INI} \\
    \midrule
    \texttt{A} & TDF + FTC & EFV & \texttt{N/A} \\
    \texttt{B} & TDF + FTC & NVP & \texttt{N/A} \\
    \texttt{C} & TDF + FTC & \texttt{N/A} & DTG \\
    \bottomrule
  \end{tabular}
  \caption{Three Examples of ART for HIV Treatments in Adolescents}
  \label{Tab:HIVExample}
\end{table}

As previously mentioned, treatment regimes for a person with HIV are changed to avoid developing drug-resistant viral strains (see \citet{bennett2008world} and Chapter 4 of \citet{world2016consolidated}). A simple scenario with three common medication combinations\footnote{See Table 4.3 on page 106 in \citet{world2016consolidated}.} for adolescents is listed in Table \ref{Tab:HIVExample} (the example medications include tenofovir (TDF), emtricitabine (FTC), efavirenz (EFV), nevirapine (NVP), and dolutegravir (DTG).). When we change the medications \texttt{OPTION A}$\rightarrow$\texttt{OPTION B}, the comp. NNRTI medication of \colorbox{yellow!25}{EFV} is replaced with the comp. NNRTI medication of \colorbox{yellow!25}{NVP}. When alternatively we change \texttt{OPTION A}$\rightarrow$\texttt{OPTION C}, \colorbox{yellow!25}{EFV} is replaced with the comp. INI medication of \colorbox{cyan!25}{DTG}.

For the examples shown in the table, sparseness arises (denoted as not applicable \texttt{N/A}) when an NNRTI medication is issued instead of an INI medication and vice versa. This also triggers \textit{strong (negative) correlations among different variables} because the existence of an NNRTI medication leads to the non-existence of an INI medication. Learning to represent multivariate categorical dependencies is thus challenging and often requires researchers to make assumptions on the dependence structure of the data \citep{chow1968approximating, dunson2009nonparametric}.

\subsection{Mode Collapse while Training GANs}\label{Sec:ModeCollapse}\label{Sec:GanDiffTrain}
There are many known practical issues for training GANs \citep{kodali2017convergence, mescheder2018training}. In particular, \textit{mode collapse} \citep{goodfellow2016nips} refers to the tendency of the GAN generator to collapse to a parametric setting where it always outputs the same point (or family of points); thus greatly reducing the diversity in GAN synthesised datasets. 

We hypothesise that mode collapse is exacerbated by the highly sparse and strongly correlated nature of multivariate categorical clinical datasets. From Table \ref{Tab:HIVExample}, we see that since \texttt{Option}s \texttt{A}, \texttt{B}, and \texttt{C} are all realistic, GAN can potentially learn to only output EFV whenever the comp. INI medication class is \texttt{N/A}. This is further complicated when the variable distributions are highly skewed (\textit{e.g.,} when EFV is prescribed more often than NVP), or when the distributions among different variables are imbalanced (\textit{e.g.,} when NNRTI medication is prescribed more often than INI medication).

Training GANs can be a difficult task. We aim to find a Nash equilibrium for the generator-discriminator pair \citep{goodfellow2014generative} which in practice is typically achieved using gradient descent techniques (\textit{e.g.,} SGD \citep{rumelhart1986learning} and Adam \citep{kingma2015adam}) rather than explicitly solving for the equilibrium strategy. Thus, training could fail to converge. Convergence in GANs can benefit from the careful selection of network modules \citep{radford2015unsupervised}, the change of learning objectives \citep{arjovsky2017wasserstein, gulrajani2017improved}, and auxiliary experimental setups \citep{salimans2016improved, sonderby2016amortised}. Nonetheless, the improvements for convergence in GAN training do not usually directly increase the quality of the generated data. As shown in \citet{gulrajani2017improved} and in \citet{liu2019spectral}, both studies found that synthetic images of higher qualities required the Lipschitz constraint to be enforced on the discriminator/critic network. However, \citet{metz2016unrolled} also argued that by dropping variety (\textit{i.e.,} by allowing mode collapse), GAN could allocate more of its expressive power to fine-tune the few and already identified modes.

One approach to mitigate mode collapse used the learnt features within the GAN sub-networks to quantify diversity in the generated data. \citet{salimans2016improved} used \textit{minibatch discrimination} to project multiple copies of generated data to a high dimensional latent space and forwarded the differences in the latent features as an extra source of information to the discriminator. \citet{li2017mmd} adopted an encoder-decoder framework for their discriminator; and the encoder compartment of their discriminator encoded both the real and synthetic data. In addition to data generation, their architectural design allowed them to train their GAN to match various levels of abstractions (\textit{i.e., moment matching}) in the encoded features of the real and synthetic data. In the same spirit, we explicitly store latent features of real data in an external buffer and replay them to the generator as a form of non-randomised prior at test time (see Figure \ref{Fig:OurDesign}).

Another line of study also showed that the generator outputs can benefit from employing discriminators with better designs or by having multiple discriminators \citep{srivastava2017veegan, mordido2020microbatchgan}. \citet{thanh2020catastrophic} found that mode collapse could be related to \textit{catastrophic forgetting} \citep{mccloskey1989catastrophic, kuo2021AAAI} in the discriminator when the discriminator parameters escaped their previous local minimum. \citet{mangalam2021overcoming} found that this could be mitigated by sequentially introducing more discriminators.

\subsection{On Concurrently Modelling Mix Typed Variables}

While \citet{goncalves2020generation} showed that the traditional GAN approach encountered difficulties in generating multivariate categorical data, real life clinical data are even more complicated and usually consist of numeric, binary, and categorical variables. Two recent studies, \citet{li2021generating} and \citet{kuo2022health}, both reported the successful generation of mixed-type datasets using GANs.

On one hand, \citeauthor{li2021generating} proposed a generator with a pair of VAEs. While VAE-GANs \citep{larsen2016autoencoding} were previously proposed to discover high-level disentangled representations in images, \citeauthor{li2021generating} used VAEs to map clinical variables of different types to a common feature representation space -- one VAE encoded the numeric variables, and the other encoded the non-numeric variables. \citeauthor{li2021generating} further included a matching loss to minimise the distance in the representation pairs.

On the other hand, \citeauthor{kuo2022health}'s Health Gym GANs included a \textit{soft-embedding} \citep{mottini2018airline} algorithm (see Figure \ref{Fig:KuoDiscriminator}). Their soft-embedding first created a small size lookup table for each binary and categorical variable; then they concatenated the lookup vectors with the numeric variables.

\subsection{The Health Gym GAN}\label{Sec:HGGAN}
We based our work on \citet{kuo2022health}'s Health Gym GAN\footnote{\label{FN:Code}See their codes in \url{https://github.com/Nic5472K/ScientificData2021_HealthGym}.}$^,$\footnote{The critic in the WGAN setting \citep{arjovsky2017wasserstein} is equivalent to the discriminator of the traditional GAN setup. It is called a critic since it’s not trained to classify outputs but to score their realisticness instead.}. Specifically, their generator $G$ with weights $\eta$ and \textit{critic} $C$ with weights $\kappa$ were trained to optimise the losses,
\begin{align}
    &L_C =  \underbrace{\mathbb{E}\left[C\big(G(z)\big)\right] - \mathbb{E}\left[C(x_{\text{real}})\right]}_{\text{Wasserstein value function}} + \underbrace{\lambda_{\text{GP}} \mathbb{E}\left[\left(\lVert{\nabla_{\Tilde{x}_{\text{syn}}}C(\Tilde{x}_{\text{syn}})}\lVert_2 - 1\right)^2\right]}_{\text{Gradient penalty loss}} \hspace{5mm}\text{and}
    \label{Eq:WGANGP_D}\\
    &L_G =-\mathbb{E}\left[ C(G(z))\right] + \underbrace{\lambda_{\text{corr}}\sum_{i = 1}^n\sum_{j = 1}^{i - 1}\text{abs}\left( r^{(i,j)}_\text{syn} - r^{(i,j)}_\text{real}\right)}_{\text{Alignment loss}},
    \label{Eq:OurGLoss}
\end{align}
following the \textit{Wasserstein GAN with gradient penalty} (WGAN-GP) setup \citep{gulrajani2017improved}. In Equation (\ref{Eq:WGANGP_D}), $G(z) = x_{\text{syn}}$ denoted the synthetic data generated from the generator with random input $z$; $x_\text{real}$ is the real data sampled from the database; $\Tilde{x}_\text{syn}$ is interpolated between $G(z)$ and $x_\text{real}$; and $\lambda_\text{GP}$ is a constant that manages the strength of the gradient penalty loss. Furthermore, \citeauthor{kuo2022health} introduced an \textit{alignment loss} in Equation (\ref{Eq:OurGLoss}) to ensure that the correlations do not diverge among variables during training. The alignment loss is computed as the sum of the absolute differences in the \textit{Pearson's r correlation} \citep{mukaka2012guide} for all pairs of distinct variables $i\neq j$ between the synthetic data $r^{(i,j)}_\text{syn}$ and their real counterparts $r^{(i,j)}_\text{real}$. A constant $\lambda_{\text{corr}}$ is employed to manage the strength of the alignment loss.

During our experiments, we found that \citeauthor{kuo2022health}'s WGAN-GP setup was able to converge without the alignment loss but could not synthesise meaningful data. Hence similar to the observation made in \citet{metz2016unrolled}, the convergence in GAN did not imply the acquisition of the ability to generate high quality output (see Section \ref{Sec:GanDiffTrain}). In contrast, the approach that we developed in this paper is capable of converging quickly and produces highly realistic data without the auxiliary objective.

\section{Materials and Methods}

We introduce an additional VAE and an extra buffer to extend \citet{kuo2022health}'s WGAN-GP setup.

\subsection{Data, Visualised from the Perspective of the Critic}\label{Sec:OverlyStructured}

\begin{wrapfigure}{r}{0.425\textwidth}
    \vspace*{-7mm}
    \centering
    \begin{subfigure}{.5\linewidth}
      \centering
      \includegraphics[width=\linewidth]{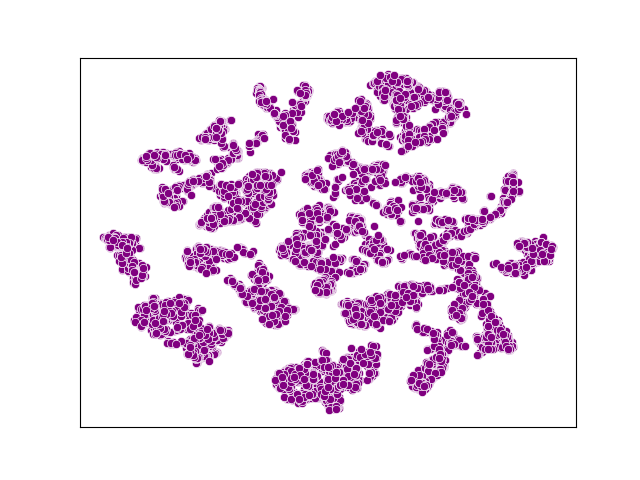}
      \caption{The $f_\text{syn}$ features.}
    \end{subfigure}%
    \begin{subfigure}{.5\linewidth}
      \centering
      \includegraphics[width=\linewidth]{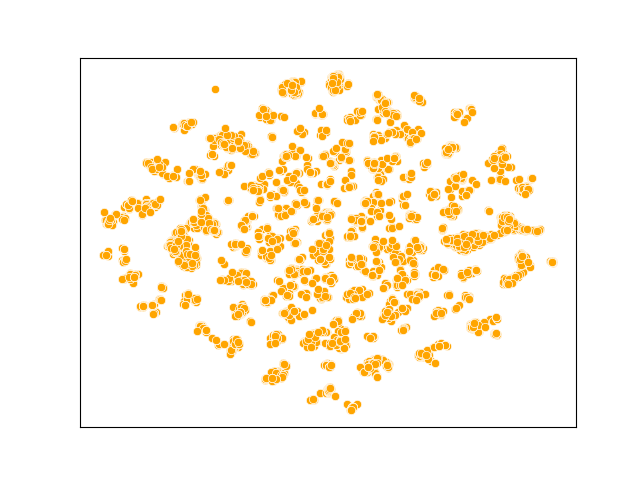}
      \caption{The $f_\text{real}$ features.}
    \end{subfigure}
    \caption{\label{Fig:tSNEFeatures}t-SNE visualisations of the\\ \hspace*{14mm}soft-embedding features.}
    \vspace*{-5mm}
\end{wrapfigure}

To illustrate the mode collapse problem, we first trained \citet{kuo2022health}'s WGAN-GP until convergence. Then, we created batches of synthetic data $G(z) = x_\text{syn}$ using random inputs $z$; and passed both $x_\text{syn}$ and the real dataset $x_\text{real}$ to their critic $C$. We extracted the features encoded in the critic's soft-embedding (a form of intermediate features, see Figure \ref{Fig:KuoDiscriminator}) as $f_\text{syn}$ and $f_\text{real}$ respectively; and then concatenated $f_\text{syn}$ and $f_\text{real}$, and used t-SNE \citep{van2008visualizing} to project the features in $\mathbb{R}^2$ and plotted them in Figure \ref{Fig:tSNEFeatures}.

The features $f_\text{syn}$ shown in subplot \ref{Fig:tSNEFeatures}(a) resembles that of a collection of clusters, while $f_\text{real}$ in subplot \ref{Fig:tSNEFeatures}(b) are much more evenly distributed in space. The overly structured $f_\text{syn}$ indicated that the generator experienced mode collapse and thus created very similar groupings of synthetic patient records and are less diversified than the real patient records.

\subsection{Feature Replay as a Form of Generator Input}\label{Sec:OurShit}

\citet{larsen2016autoencoding}'s VAE-GAN showed that smooth image interpolation could be achieved via adding attribute vectors to the learned VAE latent features. Inspired by their finding, we also extended \citet{kuo2022health}'s WGAN-GP setup with a VAE to generate variations of synthetic patient records. However, instead of sampling new latent features from the VAE learnt parameterised distributions, we externally stored and replayed the features of real data in a buffer (see Figure \ref{Fig:OurDesign}).

\begin{wrapfigure}{r}{0.6\textwidth}
    \vspace*{-5mm}
    \centering
      \includegraphics[width=\linewidth]{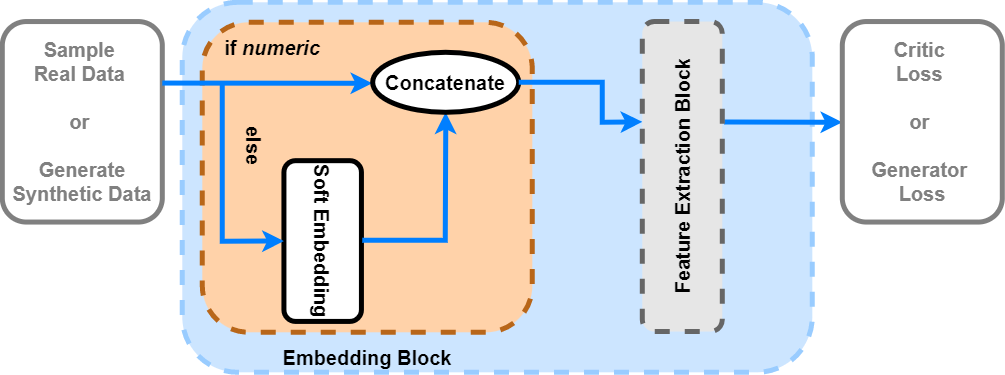}
    \caption{\label{Fig:KuoDiscriminator}The critic.}
    
\end{wrapfigure}
\textbf{Tied Soft-Embedding}\\
As shown in Figure \ref{Fig:KuoDiscriminator}, \citeauthor{kuo2022health}'s critic includes a \textit{soft-embedding} mechanism followed by a feature extraction block. Soft-embedding acts like word embedding \citep{mikolov2013distributed} and helps the critic to concurrently learn features for a mixed-type dataset.

A lookup vector is created for each available option in the binary (\textit{e.g.,} male in gender) and categorical (\textit{e.g.,} DTG in comp. INI, see Table \ref{Tab:HIVExample}) variables. Unlike the classic word embedding where all tokens shared the same embedding dimension (\textit{e.g.,} 400 dimensions per embedding for every word), soft-embedding allows each clinical variable to have a distinct dimensionality (\textit{e.g.,} 2 dimensions per class for each binary variable and 4 dimensions per class for each categorical variable). The vector representations are then concatenated with the values of the numeric variables before being further processed by the feature extraction block.

\begin{figure*}[h!]
    \centering

    \includegraphics[width=0.9\linewidth]{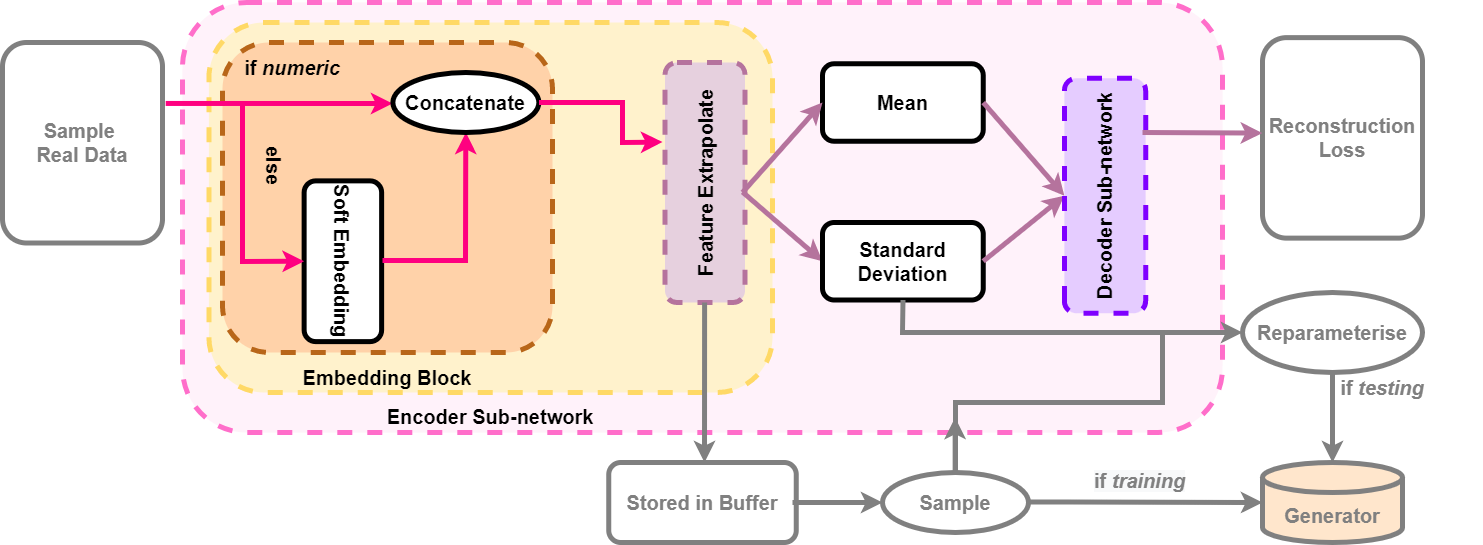}

    \caption{The VAE, with soft-embedding tied to the critic.}\label{Fig:OurEncoder}
\end{figure*}

In our extended setup, we introduce the VAE in Figure \ref{Fig:OurEncoder} with soft-embedding in its encoder. We tie the soft-embedding weights in the VAE encoder to those in the critic and hence it also observes real data. Then, we store a transformed representation $f_\text{real}$ from the real data (similar to that in Figure \ref{Fig:tSNEFeatures}(b)) in a fixed size external buffer. We replay $f_\text{real}$ as a form of non-random inputs to the generator.

\begin{algorithm}[t]
\caption{Training an Extended WGAN-GP with a VAE and a buffer.}\label{Alg:OurStuff}

\begin{algorithmic}[1]
    \STATEnonum \textbf{Initialises network parameters}: VAE's $\phi\text{ and } \theta$, as well as generator's $\eta$, and critic's $\kappa$
    \STATEnonum \textbf{Initialises an external buffer}: $\mathscr{B} \leftarrow \emptyset$

    \State \textbf{repeat}
    \STATEnonum \hspace*{5mm}\underline{\# Training the VAE $V_{\phi, \theta}$}
    \State \hspace*{5mm} 
    Sample real data 
    \hspace*{34.5mm}
    $x_\text{real} \leftarrow \mathscr{D}_\text{real}$
    \State \hspace*{5mm} 
    Apply the VAE 
    \hspace*{36.5mm}
    $\xi, \gamma_\text{real}, \sigma_V, \hat{x}_V \leftarrow V_{\phi, \theta}(x_\text{real})$
    \State \hspace*{5mm}
    Compute the VAE loss
    \hspace*{25.5mm}
    see Equation (\ref{Eq:Lv})
    \State \hspace*{5mm}
    Update the VAE parameters 
    \hspace*{18.5mm}
    $\phi, \theta \stackrel{+}\leftarrow -\nabla_{\phi, \theta}L_V$ 
    \State \hspace*{5mm}
    Release space from the buffer if full
    \hspace*{6.5mm}
    $\mathscr{B}\stackrel{\mathbb{P}_\text{release}}\leftarrow \text{cardinality of }\gamma_\text{real}$
    \State \hspace*{5mm}
    Append the buffer
    \hspace*{31.5mm}
    $\mathscr{B}\leftarrow\mathscr{B}\oplus\gamma_\text{real}$
    
    \STATEnonum\textcolor{white}{.}
    \STATEnonum \hspace*{5mm}\underline{\# Training the critic $C_\kappa$}
    \State \hspace*{5mm}
    \textbf{for }$t = 1, \ldots, n_\text{critic}$ \textbf{do}
    \State \hspace*{10mm}
    Sample real data $x_\text{real} \sim \mathscr{D}_\text{real}$, sample stored features $\gamma_\text{real} \sim \mathscr{B}$,
    \STATEnonum \hspace*{15mm}
    and sample random numbers for interpolation $\epsilon\sim U[0, 1]$
    \State \hspace*{10mm}
    Generate synthetic data 
    \hspace*{20mm}
    $x_\text{syn}\leftarrow G_\eta(\gamma_\text{real})$
    \State \hspace*{10mm}
    Syn-Real data interpolation
    \hspace*{14.25mm}
    $\Tilde{x}_\text{syn}\leftarrow \epsilon x_\text{real} + (1 - \epsilon)x_\text{syn}$
    \State \hspace*{10mm}
    Compute the critic loss
    \hspace*{20.5mm}
    see Equation (\ref{Eq:WGANGP_D})
    \State \hspace*{10mm}
    Update the critic parameters
    \hspace*{13.25mm}
    $\kappa \stackrel{+}\leftarrow -\nabla_{\kappa}L_C$
    \State \hspace*{5mm}
    \textbf{end for}
    
    \STATEnonum\textcolor{white}{.}
    \STATEnonum \hspace*{5mm}\underline{\# Training the generator $G_\eta$}
    \State \hspace*{5mm}
    Sample stored features $\gamma_\text{real} \sim \mathscr{B}$
    \State \hspace*{5mm}
    Generate synthetic data
    \hspace*{25.5mm}
    $x_\text{syn}\leftarrow G_\eta(\gamma_\text{real})$
    \State \hspace*{5mm}
    Compute the generator loss
    \hspace*{20mm}
    see Equation (\ref{Eq:OurGLoss})
    \State \hspace*{5mm}
    Update the generator parameters
    \hspace*{13mm}
    $\eta \stackrel{+}\leftarrow -\nabla_{\eta}L_G$
    
    \State \textbf{until converge}
    
\end{algorithmic}
\end{algorithm}

\textbf{Our Extra VAE}\\
A conventional VAE $V$, with encoder weights $\phi$ and decoder weights $\theta$, is trained by optimising 
\begin{align}\label{Eq:Lv}
    L_V = \underbrace{- \lambda_\text{KL}\text{KL}\left(q_\phi(\xi\lvert x_\text{real})\lVert p_\theta(\xi)\right)}_{\text{KL divergence}} + \underbrace{\mathbb{E}_{q_\phi(\xi\lvert x_\text{real})}\text{log}\left(p_\theta(x_\text{real}\lvert \xi)\right)}_{\text{Log-likelihood reconstruction loss}}.
\end{align}
$L_V$ consists of the KL divergence to approximate the posterior $q$ from the prior $p$, and a log-likelihood term to reconstruct $\hat{x}_V$ for the real data $x_\text{real}$ from the latent variable $\xi$; in addition, $\lambda_\text{KL}$ is a constant term to manage the strength of the KL divergence. 

We select a Gaussian distribution as the prior $p$. As shown in Figure \ref{Fig:OurEncoder}, our VAE encoder shares the critic's soft-embedding to encode the real data as features $\gamma_\text{real}$ and the learned standard deviations (std-s) $\sigma_V$. Then, we sample latent variables $\xi\leftarrow \gamma_\text{real} + \rho$ where $\rho\sim N(0, \sigma_V)$ for the VAE decoder to reconstruct $\hat{x}_V$.

During training, the GAN generator receives different inputs to the VAE decoder. While the VAE decoder receives latent variables $\xi$, the GAN generator receives the unmodified features $\gamma_\text{real}$ (\textit{i.e.,} $z\leftarrow \gamma_\text{real}$). In addition, we employ an external buffer $\mathscr{B}$ to collect $\gamma_\text{real}$.

At test time, we discard the VAE and sample stored features (with replacement) from the buffer for the GAN generator $\overline{\gamma_\text{real}}\sim \mathscr{B}$. The goal of the feature replay mechanic is to avoid mode collapse by establishing a dependency between the generator output and the highly diversified ground truth latent features of the real data $\gamma_\text{real}$. However, since we aim to create new synthetic patient records, the generator input are reparameterised as $z\leftarrow \overline{\gamma_\text{real}} + \rho^{*}$ where $\rho^{*}\sim N(0, \sigma_V)$.

\textbf{Our Feature Replay Mechanic \& Algorithmic Overview}\\
An overview for training our extended WGAN-GP is presented in Algorithm \ref{Alg:OurStuff}. For each iteration, we train the components in the order of: the VAE with Equation (\ref{Eq:Lv}), the critic with Equation~(\ref{Eq:WGANGP_D}) for a pre-defined inner-loop $n_\text{critic}$ \citep{gulrajani2017improved}, and the generator with Equation (\ref{Eq:OurGLoss}). Our external buffer $\mathscr{B}$ has a fixed size; when no vacancy is left in $\mathscr{B}$ (see Line 6), we randomly release space in $\mathscr{B}$ to append the new encoded features $\gamma_\text{real}$ (see Line 7). There are alternative ways to update the buffer, such as \textit{herding} \citep{welling2009herding} for constructing an exemplar set \citep{rebuffi2017icarl}; but the search for an optimal buffer update mechanism is out of the scope of this paper. 

\subsection{On VAEs: To Replay or To Sample}\label{Sec:ReplaySample}
While both our approach and \citet{larsen2016autoencoding}'s VAE-GAN employ a VAE in conjunction with a GAN, there are three important distinctions. First, \citeauthor{larsen2016autoencoding} combine the VAE decoder with the GAN generator while our setup keeps the sub-networks distinct: our VAE decoder has weights $\theta$ and the GAN generator has weights $\eta$. Second, \citeauthor{larsen2016autoencoding} implement a different reconstruction loss for the VAE. Instead of using the exact differences between the original input $x_\text{real}$ and the reconstructed outcome $\hat{x}_V$, they take the average of all differences in the features of the intermediate layers in the decoder\footnote{See Equation 7 on page 2 of \citet{larsen2016autoencoding}.}. This is similar to \citet{li2017mmd}'s moment matching practice discussed in Section \ref{Sec:GanDiffTrain}.

The third and most important difference lies in the relationship between the VAE encoder output and the generator input. In their work, the inputs of \citeauthor{larsen2016autoencoding}'s generator (merged with the decoder) are sampled from the learned distribution of the VAE encoders. In our study, we store all features $\gamma_\text{real}$ encoded from the real data and replay them to the generator (see Lines 9 and 15 in Algorithm \ref{Alg:OurStuff}). Replaying $\gamma_\text{real}$ provides exact combinations of information in the latent structure of $x_\text{real}$ to the generator and helps GAN to capture real data in the sparse feature space (see Section \ref{Sec:SparseNCorrelated}).   

\section{Experimental Setups}

This section includes the hyper-parameter settings, the baseline models, and the metrics used in our experiments.

\subsection{Hyper-Parameters}\label{Sec:Hyperparamters}

\textbf{Module Selection}\\
Our extended WGAN-GP setup inherited most of the hyper-parameters from \citet{kuo2022health}\footnote{See Fig. 1 in page 14 of \citet{kuo2022health}; and their repository in footnote \ref{FN:Code}.} including the identical designs for the generator $G$ and critic $C$. The generator had 1 bidirectional \textit{long short-term memory} (bi-LSTM) \citep{hochreiter1997long, graves2005bidirectional} followed by 3 linear layers. The critic used \colorbox{green!25}{2} and \colorbox{magenta!25}{4} as the hidden dimensions in soft-embedding for the binary and categorical variables respectively; then in the feature extraction block, it included 2 linear layers followed by 1 bi-LSTM and 1 additional linear layer. The input and hidden dimensions of $G$ were 128. The hidden dimension for $C$ was also 128; and the input dimension was 29 (from concatenating the soft-embeddings of 3 numeric, 5 binary, and 4 categorical variables $3 + 5 \times \colorbox{green!25}{2} + 4 \times \colorbox{magenta!25}{4} = 29$).

The encoder in the VAE $V$ shared the soft-embedding in the critic. The encoder then employed 4 linear layers with \textit{residual connections} \citep{he2016deep} between each layer for feature extrapolation. The decoder of $V$ had 1 linear layer. The hidden dimension of $V$ was 128 hence $\gamma_\text{real}, \sigma_V\in\mathbb{R}^{128}$ (matching the input dimension of the generator). Since the VAE shared the soft-embedding of $C$, $V$'s input dimension was 29.

The latent features encoded from real data were stored in a fixed size buffer (see Lines 6 and 7 in Algorithm \ref{Alg:OurStuff}). We defaulted the buffer size to 10,000 samples, storing roughly 3\% of all real data. However, we also tested when less memory was allocated.

\textbf{Optimisation Setup}\\
The sub-networks $G$, $C$, and $V$ were all trained using Adam \citep{kingma2015adam} with learning rate $0.001$ for 200 epochs with batch size 256. We set the regularisation coefficients in GANs as $\lambda_\text{GP} = 10$ and $\lambda_\text{corr} = 10$ (see Equations (\ref{Eq:WGANGP_D}) and (\ref{Eq:OurGLoss})); and $\lambda_\text{KL} = 1$ (see Equation (\ref{Eq:Lv})) for the VAE. In addition, we updated $C$ for $n_\text{critic} = 5$ steps  (see Line 8 of Algorithm \ref{Alg:OurStuff}) for every update in $G$. 

As we previously discussed in Section \ref{Sec:HGGAN}, the alignment loss regularised with $\lambda_\text{corr}$ was introduced in \citet{kuo2022health} to stabilise GAN training. In our extended WGAN-GP training we included the alignment loss; but we also tested performance without the auxiliary loss. Furthermore, the medication records of ART for HIV have non-uniform lengths (see Section \ref{Sec:GroundTruth}). The shortest record was 10 units long and the longest record was 100 units long; hence we employed \textit{curriculum learning} \citep{bengio2009curriculum} to sequentially expose our models to longer records.  

\subsection{Baseline Models}\label{Sec:Baselines}
We refer to \citet{kuo2022health}'s setup as \textcolor{brown}{\texttt{WGAN-GP}} and ours as \textcolor{magenta}{\texttt{WGAN-GP+VAE+Buffer}}.

\textbf{Alternative Architectural Designs}\\
While the original \textcolor{brown}{\texttt{WGAN-GP}} reported competitive results, it could potentially benefit from being equipped with alternative modules that were developed more recently. Both the critic and the generator of \textcolor{brown}{\texttt{WGAN-GP}} employed LSTMs (\citet{hochreiter1997long}, see Section \ref{Sec:Hyperparamters}). We tested various ways to substitute their LSTMs with -- Transformer \citep{vaswani2017attention}, BERT-like encoder-only Transformer \citep{kenton2019bert}, and GPT-like decoder-only Transformer \citep{radford2018improving}. We found that the best combination resulted from replacing the bi-LSTM in the generator with encoder-only Transformers (EOTs). We implemented two versions of GAN with EOT:\\ \hspace*{5mm}\texttt{WGAN-GP+G$\_$EOT} with EOT na\"ively implemented; and\\ 
\hspace*{5mm}\textcolor{cyan}{\texttt{WGAN-GP+G$\_$EOT+VAE+Buffer}} when EOT was implemented along with our extended setup.

Our EOT modules processed input data sequentially with 3 sets of multi-head scaled dot-product attention. Following the setting in Section \ref{Sec:Hyperparamters}, the input and hidden dimensions were 128 and we employed 8 heads. Importantly, attention mechanism was applied to the temporal dimension\footnote{Attention mechanism was not applied to the feature dimension (\textit{i.e.,} the clinical variables in Table \ref{Tab:VarsOfHIV}). This was because in their paper, \citet{kuo2022health} found that their simpler LSTM-based GAN was already capable of capturing correlations between clinical variables.}. In addition, there were 100 lookup vectors in the EOT positional embedding; this was because the longest ART for HIV record was 100 units long.

We tested \citet{larsen2016autoencoding}'s VAE-GAN against our \texttt{WGAN-GP+G$\_$EOT}; see Section \ref{Sec:ReplaySample} for an in-depth comparison of the two models. A VAE encoder was added on top of \citeauthor{kuo2022health}'s original design (with gradient penalty included) and we referred to it as \texttt{VAE-WGAN-GP}. 

\textbf{Previous Methods that Aim to Mitigate Mode Collapse}\\
We further extended \textcolor{brown}{\texttt{WGAN-GP}} with 3 techniques that were initially proposed to mitigate the mode collapse of GANs for image generation (see Section \ref{Sec:GanDiffTrain}): \texttt{WGAN-GP+MBD} with minibatch discrimination (MBD) \citep{salimans2016improved}, \texttt{WGAN-GP+MM} with moment matching (MM) \citep{li2017mmd}, and \texttt{WGAN-GP+MC} with multiple critics (MC) \citep{mangalam2021overcoming}. Specifically, \texttt{WGAN-GP+MBD} was implemented with projection matrices of size $\mathbb{R}^{3\times5}$ after the embedding block and before the feature extraction block of the critic; \texttt{WGAN-GP+MM} matched the features encoded by the critic embedding block; and we introduced a new critic every 50 epochs for \texttt{WGAN-GP+MC}. 

\subsection{Metrics}\label{Sec:Metrics}
There are 5 desiderata: $\mathfrak{a}$) that our technique mitigates mode collapse in the GAN generator; $\mathfrak{b}$) that all generated variables are individually realistic; $\mathfrak{c}$) that all variables are also collectively realistic over time; $\mathfrak{d}$)  that our synthetic datasets are highly secure and do not disclose patient identity; and last $\mathfrak{e}$) that our dataset has high utility and can substitute a real dataset for downstream model building.

\subsubsection{Mitigating Mode Collapse}\label{Sec:Metric01}

Following \citet{goncalves2020generation}, we employ the \textit{log-cluster} metric $U$ \citep{woo2009global}
\begin{align}
    U = \text{log}\left(\frac{1}{\Gamma}\sum_{k = 1}^\Gamma\left[\frac{n_{k_\text{real}}}{n_{k}} - \frac{n_{k_\text{real}}}{n_{k_\text{real}} + n_{k_\text{syn}}}\right]^2\right)
\end{align}
to estimate the difference in latent structures between the real and the synthetic datasets. First, we sample records from both the real and synthetic datasets. Then, we merge the sub-datasets and perform a cluster analysis via k-means with $\Gamma = 20$ clusters. We denote $n_k$ as the total amount of records in cluster $k$; and $n_{k_\text{real}}$ and $n_{k_\text{syn}}$ are the respective number of real and synthetic records in the cluster. $U$ thus measures the logged average cluster-wise divergence. We repeat this process for 20 times for each synthetic dataset; and each time we sample 100,000 real and synthetic records. The lower the $U$ score, the more realistic the synthetic datasets.

In addition to the log-cluster metric, we also include the \textit{category coverage} (CAT) metric
\begin{align}
    \text{CAT} = \frac{1}{\mathscr{U}}\sum_{u = 1}^{\mathscr{U}}\frac{\lVert \mathfrak{D}^{(u)}_{\text{syn}} \lVert}{\lVert \mathfrak{D}^{(u)}_{\text{real}} \lVert}
\end{align}
proposed in \citet{goncalves2020generation}. For all binary and categorical variables $u = 1,\ldots, \mathscr{U}$, we find the number of unique levels $\lVert\cdot\lVert$ of the $u$-th variable in the synthetic dataset $\mathfrak{D}_\text{syn}$ and its real counterpart $\mathfrak{D}_\text{real}$. The higher the CAT score (CAT$\in[0, 1]$), the more complete the non-numeric classes in the synthetic datasets and the less likelihood of mode collapse in the GAN generators.

While convergence does not directly guarantee the generation of high quality synthetic datasets (see Section \ref{Sec:GanDiffTrain}); GANs that do not converge are unlikely to generate good results. Our convergence metric is based on the correlation loss $L_{\text{corr}}=\sum_{i = 1}^n\sum_{j = 1}^{i - 1}\text{abs}\left( r^{(i,j)}_\text{syn} - r^{(i,j)}_\text{real}\right)$ of Equation (\ref{Eq:OurGLoss}); and we record this score for every iteration during the GAN training phase. This allows us to know whether the models are learning; and whether the relations among the clinical variables are learned. The lower the $L_{\text{corr}}$ score, the better represented the correlations among the synthetic data variables.

\subsubsection{Realisticness of the Individual Variables}\label{Sec:RealInd}
The individual realisticness of each variable can be checked with 2 types of plots. For the numeric variables, we superimpose the synthetic distribution over their real counterparts using \textit{kernel density estimations} (KDEs) \citep{davis2011remarks}. As for the binary and categorical variables, we show the percentage share of each option using side-by-side barplots. 

\begin{wrapfigure}{r}{0.4\textwidth}
    \centering
    \vspace*{-2mm}
    \includegraphics[width=\linewidth]{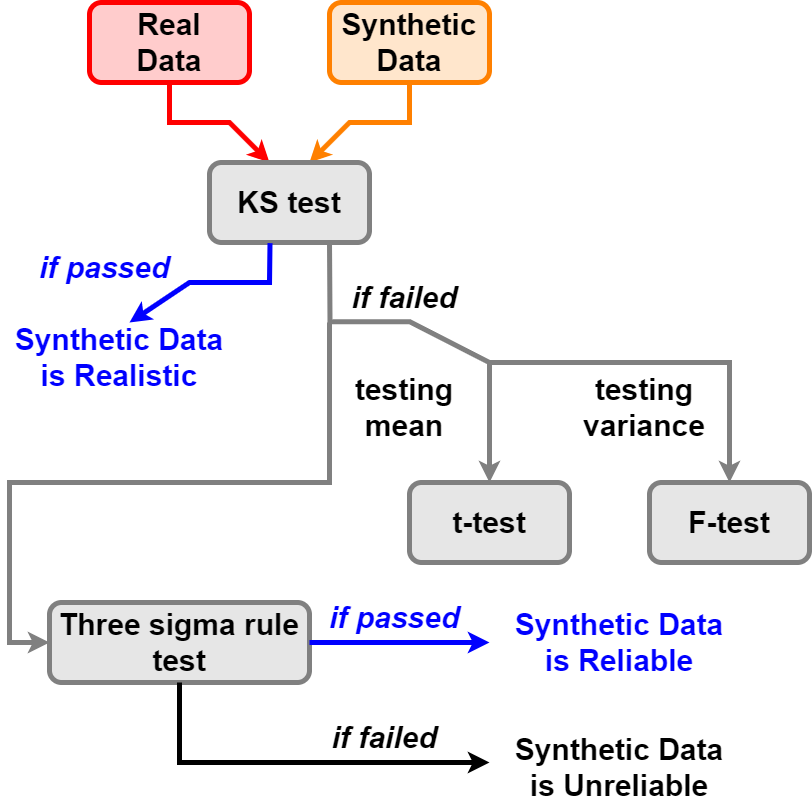}
    \caption{\label{Fig:HST}Statistical tests.}
    \vspace*{-3mm}
\end{wrapfigure}

Following \citet{kuo2022health}, we perform 4 statistical tests on the synthetic datasets organised as shown in Figure \ref{Fig:HST}. We start with the two-sample Kolmogorov–Smirnov (KS) test \citep{hodges1958significance} to check whether the synthetic variables capture details of the distributions of their real counterparts. If a synthetic variable passes the KS test, the synthetic data is realistic and is considered to have been drawn from the real database. If a synthetic variable fails the KS test, we also aim to understand why it is considered non-realistic.

The two independent sample Student's t-test \citep{yuen1974two} and F-test (numeric: Snedecor’s F-test \citep{snedecor1989statistical}; otherwise: the analysis of variance F-test) are useful techniques to understand why a synthetic variable would be considered unrealistic. The former checks the alignment between the synthetic and real mean; while the latter verify the alignment with the synthetic and real variance. 

While the t-test and F-test could help us identify the shortcomings of a synthetic variable, they cannot assess the reliability of a synthetic variable if it fails the KS test. To this end, we employ the three sigma rule test \citep{pukelsheim1994three} (defaulted with 2 std-s) to check if the values in the synthetic variable fall in a probable range of real variable values.  

\subsubsection{Fidelity in Variable Correlations}\label{Sec:FidelCorr}
We compute Kendall's $\tau$ rank correlation \citep{kendall1945treatment} to compare the relationships of the variables in the mixed-type datasets\footnote{Unlike Pearson's correlation \citep{mukaka2012guide}, Kendall's correlation can be applied to both numeric and non-numeric variables; and also between a numeric and non-numeric pair of variables.}. The correlation is computed in 2 different ways. First, we compute the \textit{static} correlations among all data points under the classic setup. Second, we follow the practice in \citet{kuo2022health} and estimate the average \textit{dynamic} correlations.

The dynamic correlation is computed in two parts. First, we take all records of a patient $\mathfrak{p}$ and linearly deconstruct each variable $i$ into a trend with cycle
\begin{align}
    X^{(i)}_{\mathfrak{p}} &= \text{Trend}(X^{(i)}_\mathfrak{p}) + \text{Cycle}(X^{(i)}_\mathfrak{p}).
\end{align}
Trends describe macroscopic behaviours in the time series data (\textit{i.e.,} increasing or decreasing over time); and cycles helps us understand information on the microscopic level (\textit{e.g.,} periodic behaviours). After detrending the variables, we compute the correlations in both the trends and cycles; and then we average the values across all patients.

\subsubsection{Patient Disclosure Risk}\label{Sec:PDR}
We conduct two tests to check whether our synthetic datasets are highly secure for public access. First, we verify that the minimum Euclidean distance between all synthetic records and real records is greater than zero thus no real record is leaked into the synthetic datasets. Then, we use \citet{el2020evaluating}'s \textit{sample-to-population attack} to check if an attacker can learn new information by matching an individual from the synthetic datasets to the real datasets.

\citeauthor{el2020evaluating} estimated the patient exposure risk via \textit{quasi-identifiers} -- special variables that may reveal the identity of a person. The quasi-identifiers in the ART for HIV dataset are \textit{Gender} and \textit{Ethnicity}. Equivalent classes are made from combinations of quasi-identifiers (\textit{e.g.,} \textit{Male + Asian} and \textit{Female + African}). Then, for every synthetic patients $s$, we compute
\begin{align}
    \frac{1}{S}\sum_{s = 1}^{S}\left(\frac{1}{F_{s}} \times I_{s}\right)
    \label{Eq:FakeRisk}
\end{align}
to estimate the chance of matching a random individual in the synthetic dataset to an individual in the real dataset. $S$ is the total records in the synthetic dataset, $I_{s} \in \{0, 1\}$ equals to one if the equivalent class of synthetic $s$ is present in both datasets (\textit{e.g.,} \textit{Male + Asian} appears in both the real and synthetic datasets), and $F_{s}$ is the cardinality of the equivalent class in the real dataset (\textit{i.e.,} the target dataset). The \citet{european2014european} and \citet{canadian2019canadian} have both suggested that this risk should be no more than $9\%$ to balance security and synthetic data utility\footnote{For the \citet{european2014european}, refer to Sec. 5.4 clause 4 on page 47. For \citet{canadian2019canadian}, refer to Sec. 6.2 under the Risk Threshold subsection.}. 

\subsubsection{Utility Verification}\label{Sec:Uti01}
We compare RL agents trained on both the real and synthetic datasets. Then, we claim that a synthetic dataset has a high level of utility if the two trained RL agents suggest similar actions to treat patients; implying that the synthetic dataset can be used in place of the real dataset to support the development of downstream machine learning applications.

We define a set of observational variables $\mathcal{D}_O$ and a set of action variables $\mathcal{D}_A$ from Table \ref{Tab:VarsOfHIV} to train our RL agents. In our experiments, $\mathcal{D}_O$ comprises the three numeric variables -- VL, CD4, and Rel CD4 -- at the current time step $t$; and all of the medications used in the previous time step $t - 1$ (five variables, as described in Section \ref{Sec:GroundTruth}). The three numeric variables inform us of a patient's health state, and we factor information of previous regimens to consider the likelihood of formation of drug-resistant viral strains (see Section \ref{Sec:SparseNCorrelated}). 

We follow the work of \citet{liu2021offline} to define clinical states from the observation variables $\mathcal{D}_O$. We first apply cross decomposition \citep{Wegelin00asurvey} to reduce the dimensionality of $\mathcal{D}_O$ to 5. Then, we perform K-Means clustering \citep{vassilvitskii2006k} with 100 clusters to label each data point in $\mathcal{D}_O$ using their associated clusters.

While the observation variables always consist of the same set of variables, we compare various setups for defining the action variables $\mathcal{D}_A$. In one of our setups, we reserve Base Drug Combo and Comp. NNRTI to span the action space $A$, resulting in $24$ ($=6\times4$) unique actions. Refer to Section \ref{Sec:Uti02Z} for all of our setups.

Our RL method of choice is batch-constrained Q-learning \citep{fujimoto2019off}. In each setup, we update the policy for 100 iterations with step size 0.01. Several alternative offline RL methods \citep{levine2020offline} have recently been developed; but we employ RL in this paper to verify synthetic dataset utility, and hence a large scale comparison over different RL techniques is out of the scope of this study. 

We update the RL policy using the reward function adapted from \citet{parbhoo2017combining}. The reward function is given as
\begin{align}\label{Eq:HivReward}
    \texttt{reward}_t = \left\{
            \begin{array}{ll}
                  -0.7 \textcolor{white}{..} \text{log}\texttt{VL}_t + 0.6 \textcolor{white}{..} \text{log} \texttt{CD4}_t, &\text{ if }\texttt{VL}_t\text{ is above detection limits, and}\\
                  5 + 0.6 \textcolor{white}{..} \text{log}\texttt{VL}_t &\text{ if }\texttt{VL}_t\text{ is below detection limits}. 
            \end{array} 
        \right.
\end{align}
Note, we use cells/$\mu L$ as the unit for $\texttt{CD4}_t$ count in Table \ref{Tab:VarsOfHIV}, while \citeauthor{parbhoo2017combining} use cells/mL as their unit for the reward computation in Equation (\ref{Eq:HivReward}).

\newpage
\section{Results}\label{Sec:ResultsHIV}
This section evaluates the five desiderata outlined in Section \ref{Sec:Metrics}.

\subsection{Mode Collapse and Training Stability}\label{Sec:MCTS}
\begin{table}[ht!]
\centering
\caption{\label{Tab:Metrics_HIV}Metric comparisons. $\downarrow$: the lower the better; $\uparrow$: the higher the better.}
\scriptsize
\begin{tabular}{llll}\toprule  
& Logged correlation ($L_\text{corr}$) ($\downarrow$)  & Log-cluster ($U$) ($\downarrow$) & Category coverage (CAT) ($\uparrow$) \\\midrule
\textcolor{brown}{\texttt{WGAN-GP}} \citep{kuo2022health} & 
$-7.91$ & $-2.11 \pm 0.02$ & $0.98$\\  \midrule\midrule
\textcolor{magenta}{\texttt{WGAN-GP+VAE+Buffer}} (Ours) & 
$\mathbf{-8.09}$ & $-2.65 \pm 0.05$ & $\mathbf{1.00}$\\  \midrule
\textcolor{cyan}{\texttt{WGAN-GP+G\_EOT+VAE+Buffer}} (Ours) & 
$-7.97$ & $\mathbf{-3.00} \pm 0.02$ & $\mathbf{1.00}$\\  \midrule\midrule
\texttt{VAE-WGAN-GP} \citep{larsen2016autoencoding} & 
$-7.86$ & $-2.24 \pm 0.04$ & $0.93$\\  \midrule
\texttt{WGAN-GP+MBD} \citep{salimans2016improved} & 
$-7.32$ & $-1.79 \pm 0.04$ & $0.90$\\  \midrule
\texttt{WGAN-GP+MM} \citep{li2017mmd} & 
$-7.49$ & $-2.31 \pm 0.03$ & $0.95$\\  \midrule
\texttt{WGAN-GP+MC} \citep{mangalam2021overcoming} & 
$-7.93$ & $-2.27 \pm 0.03$ & $0.93$\\  \bottomrule
\end{tabular}
\end{table}

\begin{figure}[ht!]
    \centering
    \begin{subfigure}{0.475\linewidth}
      \centering
      \includegraphics[width=\linewidth]{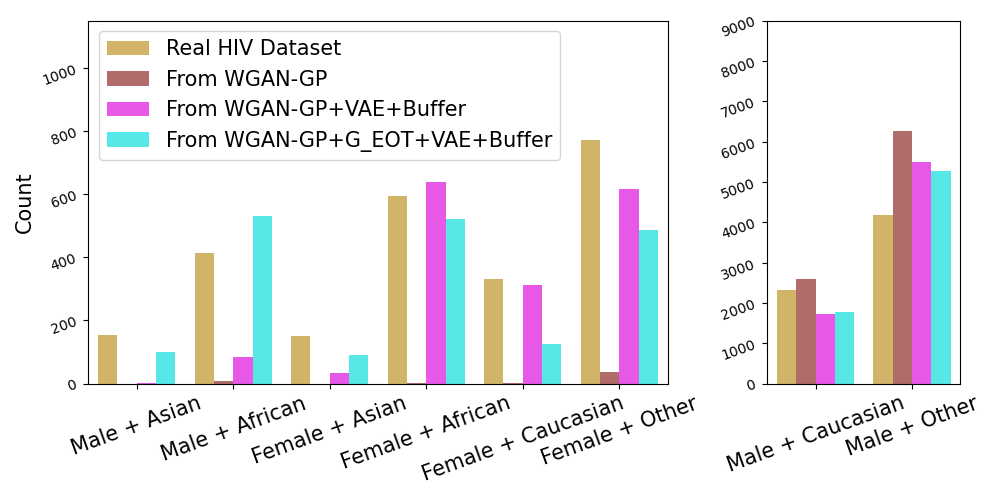}
      \caption{Real dataset vs our synthetic dataset candidates.}
    \end{subfigure}
    \begin{subfigure}{0.475\linewidth}
      \centering
      \includegraphics[width=\linewidth]{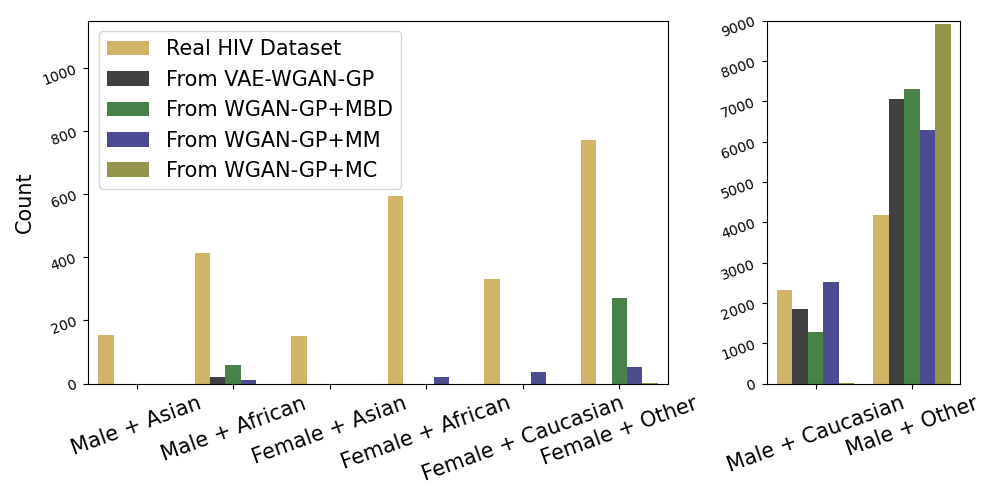}
      \caption{Other synthetic dataset candidates.}
    \end{subfigure}%
    
    \caption{\label{Fig:EthGen}Inspecting mode collapse on the gender-ethnicity pair.}
\end{figure}

\subsubsection{On Mitigating Mode Collapse}
After training seven variants of GANs for generating ART for HIV data, we compared their synthetic datasets using the metrics discussed in Section \ref{Sec:Metric01} and aggregated the results in Table \ref{Tab:Metrics_HIV}. We found that our extra VAE with external buffer performed more favourably in mitigating mode collapse. Specifically, the lower log-cluster scores indicate that our synthetic datasets $\mathfrak{D}_\text{alt}$ are better at mirroring the latent structure in the ground truth $\mathfrak{D}_\text{real}$ (refer to the denotations in Section \ref{Sec:Introduction}). In addition, both of our $\mathfrak{D}_\text{alt}$ score $1.00$ in category coverage, thus they cover all categories that can be found in $\mathfrak{D}_\text{real}$. While these quantitative analyses provided an easy way to quickly compare multiple GAN designs, these metrics remain purely relative and do not carry any physical meaning. We hence conducted a qualitative study to investigated the impact of mode collapse.

We plotted the frequency of all \textit{gender-ethnicity pairs} in Figure \ref{Fig:EthGen}. Subplot \ref{Fig:EthGen}(a) showed that the synthetic dataset $\mathfrak{D}_\text{null}$ generated using \citet{kuo2022health}'s \textcolor{brown}{\texttt{WGAN-GP}} indicated mode collapse and created synthetic patients of mostly \textit{Male+Caucasian} and \textit{Male+Other}. In contrast, our synthetic datasets $\mathfrak{D}_\text{alt}$ generated from \textcolor{magenta}{\texttt{WGAN-GP+VAE+Buffer}} and \textcolor{cyan}{\texttt{WGAN-GP+G\_EOT+VAE+Buffer}} captured both genders for all ethnicity. 

We also found that previous methods that were proposed to mitigate mode collapse in computer vision were not very effective for clinical time-series. In subplot \ref{Fig:EthGen}(b), the synthetic datasets generated using MBD \citep{salimans2016improved}, MM \citep{li2017mmd}, and MC \citep{mangalam2021overcoming} all behaved similarly to (or worse than) \citeauthor{kuo2022health}'s baseline. These techniques were likely ineffective due to the unique challenges in real life clinical datasets previously mentioned in Section \ref{Sec:SparseNCorrelated}.

Figure \ref{Fig:EthGen} also showed that mode collapse in GANs would decrease cohort diversity through omitting patients of minority background (\textit{e.g.,} Asian males and African females). This meant that if we were to ignore mode collapse and use the synthetic dataset $\mathfrak{D}_\text{null}$ to develop downstream RL agents, we would not be able to suggest optimal treatments for under-represented patients. See more results in Section \ref{Sec:Uti02Z} regarding utility verification. 

\subsubsection{Ablation Study on Replay Diversity}
\begin{wrapfigure}{r}{0.475\textwidth}
    \vspace*{-5mm}
    \centering
    \includegraphics[width=\linewidth]{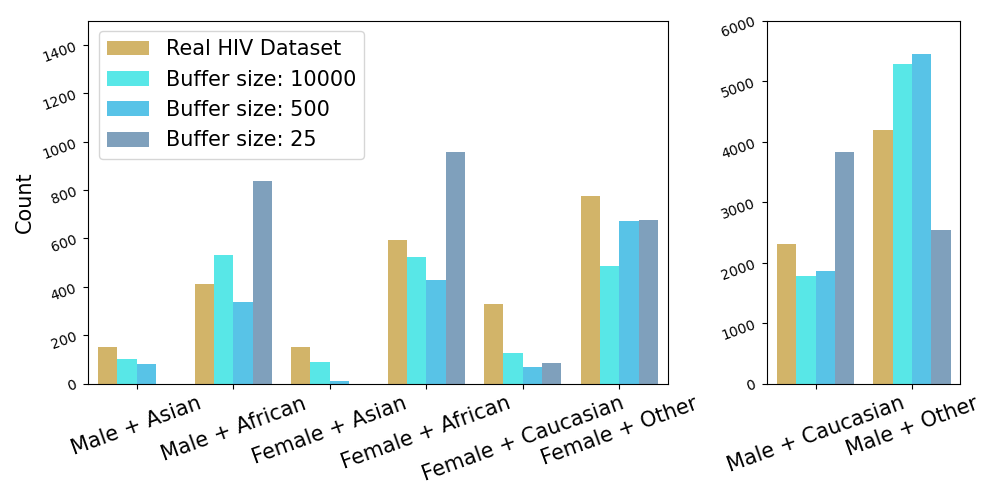}
    \caption{\label{Fig:DiffBuff}Inspecting buffers of different sizes.}
    \vspace*{-3mm}
\end{wrapfigure}
Overall, our \textcolor{cyan}{\texttt{WGAN-GP+G\_EOT+VAE+Buffer}} performed most favourably across our metrics. However, the success of this setup came with the extra memory cost from the use of the external buffer. In order to test the robustness of our novel setup, we reduced the buffer from a large size of 10,000 samples to a small size of 500 samples; and to a tiny size of 25 samples. The large buffer stored $3.00\%$ of HIV records, and the small and tiny buffers stored $0.15\%$ and $0.008\%$ of records, respectively\footnote{See Section \ref{Sec:GroundTruth}, the real HIV dataset contains $332,800$ records in total.}.

The test results in Figure \ref{Fig:DiffBuff} show that the larger the buffer size the higher the quality of the synthetic dataset. Both the large and small buffers helped our setup to generate patients of all demographics, but the tiny buffer had difficulties in creating Asian patient records. Note, the tiny buffer size still produced a synthetic dataset of higher quality than \citet{kuo2022health}'s \textcolor{brown}{\texttt{WGAN-GP}} (see Figure \ref{Fig:EthGen}(a)).

\subsubsection{Training Stability}
\begin{figure}[ht!]
    \centering
    \begin{subfigure}{.33\linewidth}
      \centering
      \includegraphics[width=\linewidth]{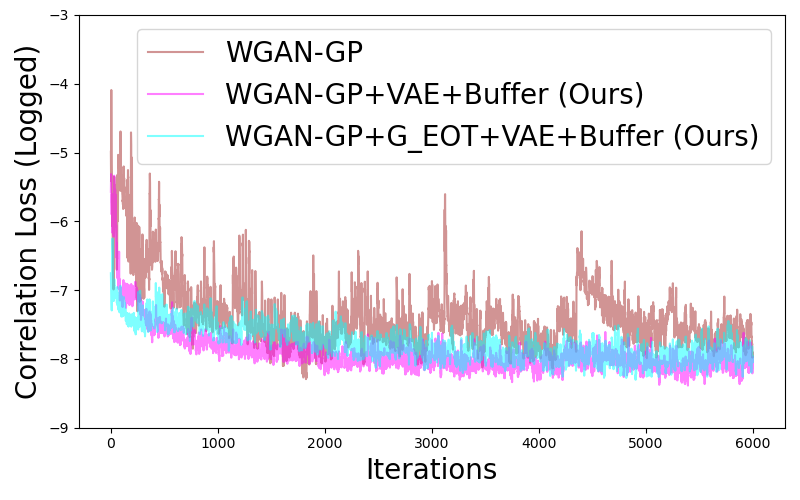}
      \caption{Ours vs \citet{kuo2022health}.}
    \end{subfigure}%
    \begin{subfigure}{.29\linewidth}
      \centering
      \includegraphics[width=\linewidth]{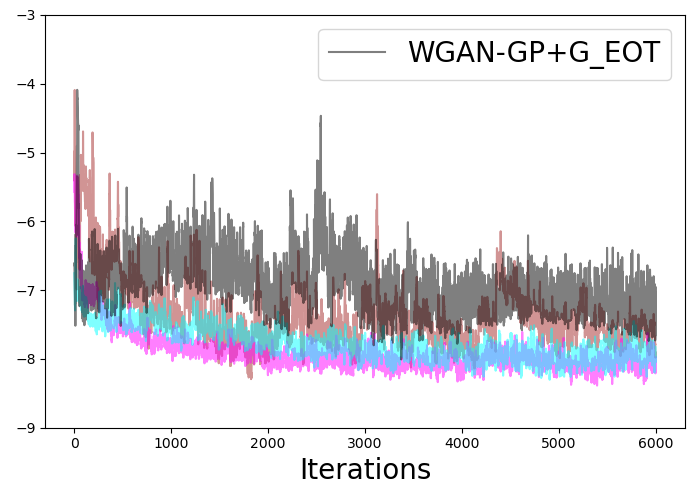}
      \caption{Na\"ive Transformers.}
    \end{subfigure}
    \begin{subfigure}{.29\linewidth}
      \centering
      \includegraphics[width=\linewidth]{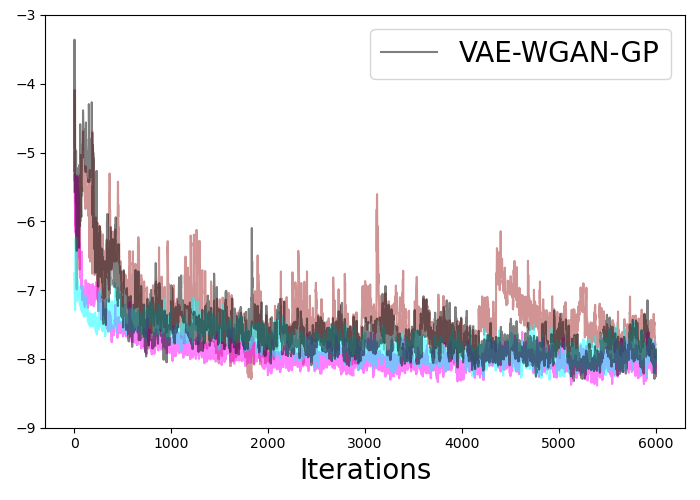}
      \caption{Idea of \citet{larsen2016autoencoding}.}
    \end{subfigure}
    
    \begin{subfigure}{.33\linewidth}
      \centering
      \includegraphics[width=\linewidth]{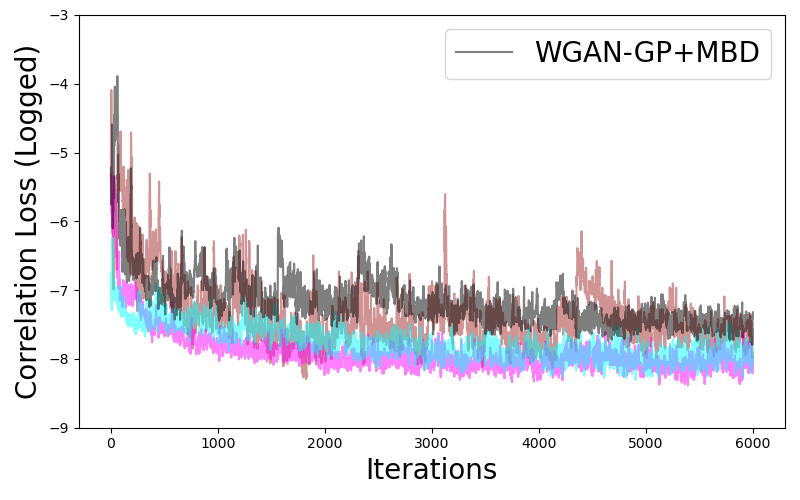}
      \caption{Idea of \citet{salimans2016improved}.}
    \end{subfigure}%
    \begin{subfigure}{.29\linewidth}
      \centering
      \includegraphics[width=\linewidth]{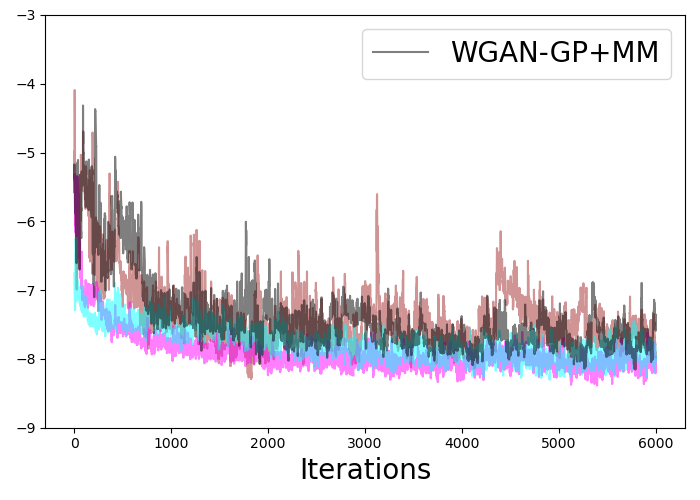}
      \caption{Idea of \citet{li2017mmd}.}
    \end{subfigure}
    \begin{subfigure}{.29\linewidth}
      \centering
      \includegraphics[width=\linewidth]{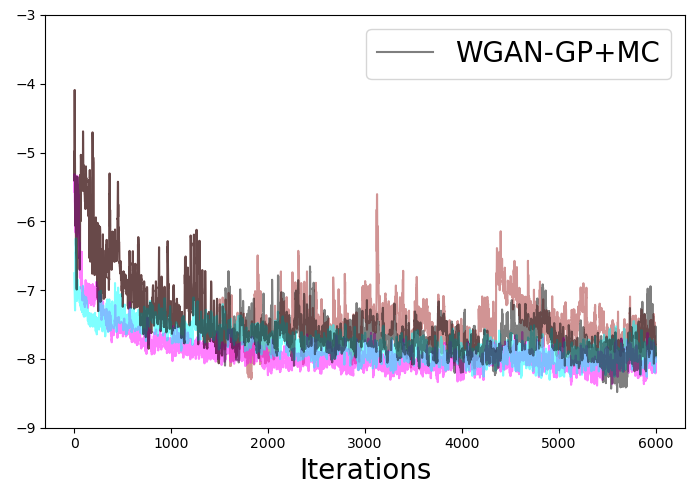}
      \caption{Idea of \textcolor{blue}{Mangalam} (\citeyear{mangalam2021overcoming}).}
    \end{subfigure}
    
    \caption{\label{Fig:HIVCurves}The logged correlation loss of the first 200 epochs.}
\end{figure} 

As discussed in Section \ref{Sec:GanDiffTrain}, the instability experienced by GANs during training is highly entwined with mode collapse. In Figure \ref{Fig:HIVCurves}, we illustrated the correlation loss $L_\text{corr}$ during the training phase of different GAN variants; and we observed in subplot \ref{Fig:HIVCurves}(a) that the additional VAE and extra buffer in our \textcolor{magenta}{\texttt{WGAN-GP+VAE+Buffer}} and \textcolor{cyan}{\texttt{WGAN-GP+G\_EOT+VAE+Buffer}} were beneficial for achieving lower loss in less time. While the other baseline models also converged, their optimisation were not as smooth and Table \ref{Tab:Metrics_HIV} showed that they scored worse in the final $L_\text{corr}$ metric.

Some interesting observations were made when we replaced the LSTMs with Transformers in the GAN generator. Subplot \ref{Fig:HIVCurves}(b) demonstrates that a na\"ive replacement would lead to worse results; and subplot \ref{Fig:HIVCurves}(a) shows that \textit{Transformers could perform more favourably than LSTMs when implemented along with our extended replay mechanism}. This is likely because Transformer's attention mechanism \citep{bahdanau2015neural} acted as a form of weighted aggregation. Under the classic GAN setup, randomly sampled vectors were forwarded to the Transformer in the GAN generator as inputs, and attention weights assigned to such random vectors were not meaningful. In contrast, our replay algorithm supplied the Transformer in the GAN generator with information on the latent structure of the real data; Transformer's attention mechanism was hence able to extract meaningful information to enhance the quality of synthetic data.

\begin{figure}[t]
    \centering
    \begin{subfigure}{.33\linewidth}
      \centering
      \includegraphics[width=\linewidth]{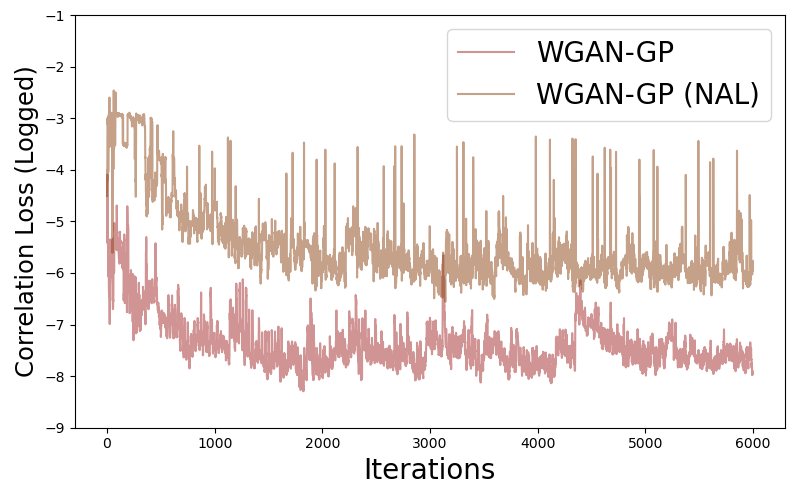}
      \caption{The classic WGAN-GP.}
    \end{subfigure}%
    \begin{subfigure}{.29\linewidth}
      \centering
      \includegraphics[width=\linewidth]{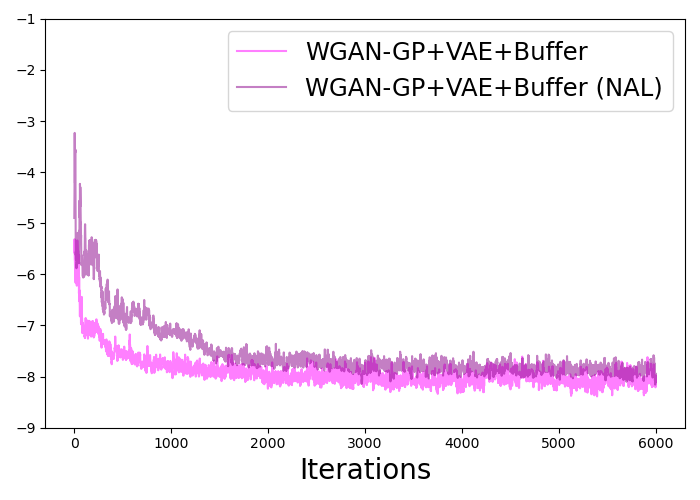}
      \caption{Our setup.}
    \end{subfigure}
    \begin{subfigure}{.29\linewidth}
      \centering
      \includegraphics[width=\linewidth]{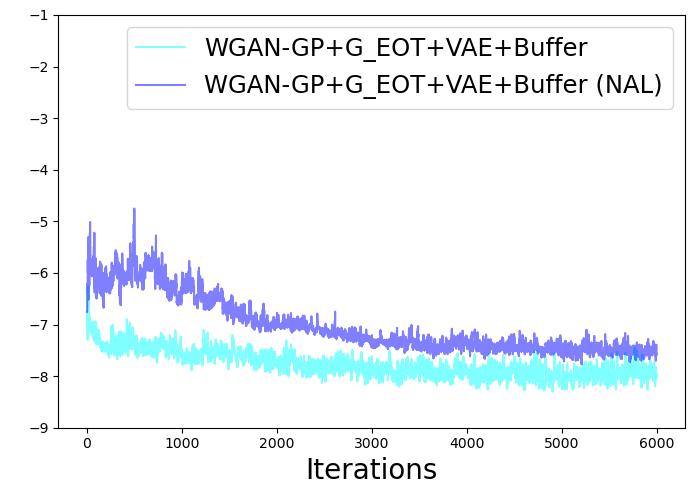}
      \caption{Our setup with EOT.}
    \end{subfigure}

    \caption{\label{Fig:NAL}Training with and with no alignment loss (NAL).}
\end{figure}

We further tested the training stability of GANs with no alignment loss (NAL) (see Equation (\ref{Eq:OurGLoss})). As shown in Figure \ref{Fig:NAL}, the alignment loss was essential for updating \citet{kuo2022health}'s \textcolor{brown}{\texttt{WGAN-GP}}. Without this auxiliary learning objective, \textcolor{brown}{\texttt{WGAN-GP}} converged to a larger minimum loss indicating that its generated data was less realistic. In contrast, our extended GAN optimised successfully under the NAL scenario; no explicit assumptions on the data structure was therefore required for generating our synthetic data. Nonetheless, training was faster with the alignment loss employed. 

\subsection{Realisticness of the Individual Variables}\label{Sec:RIV}

\begin{figure}[ht!]
    \centering
    \begin{subfigure}{0.825\linewidth}
      \centering
      \includegraphics[width=\linewidth]{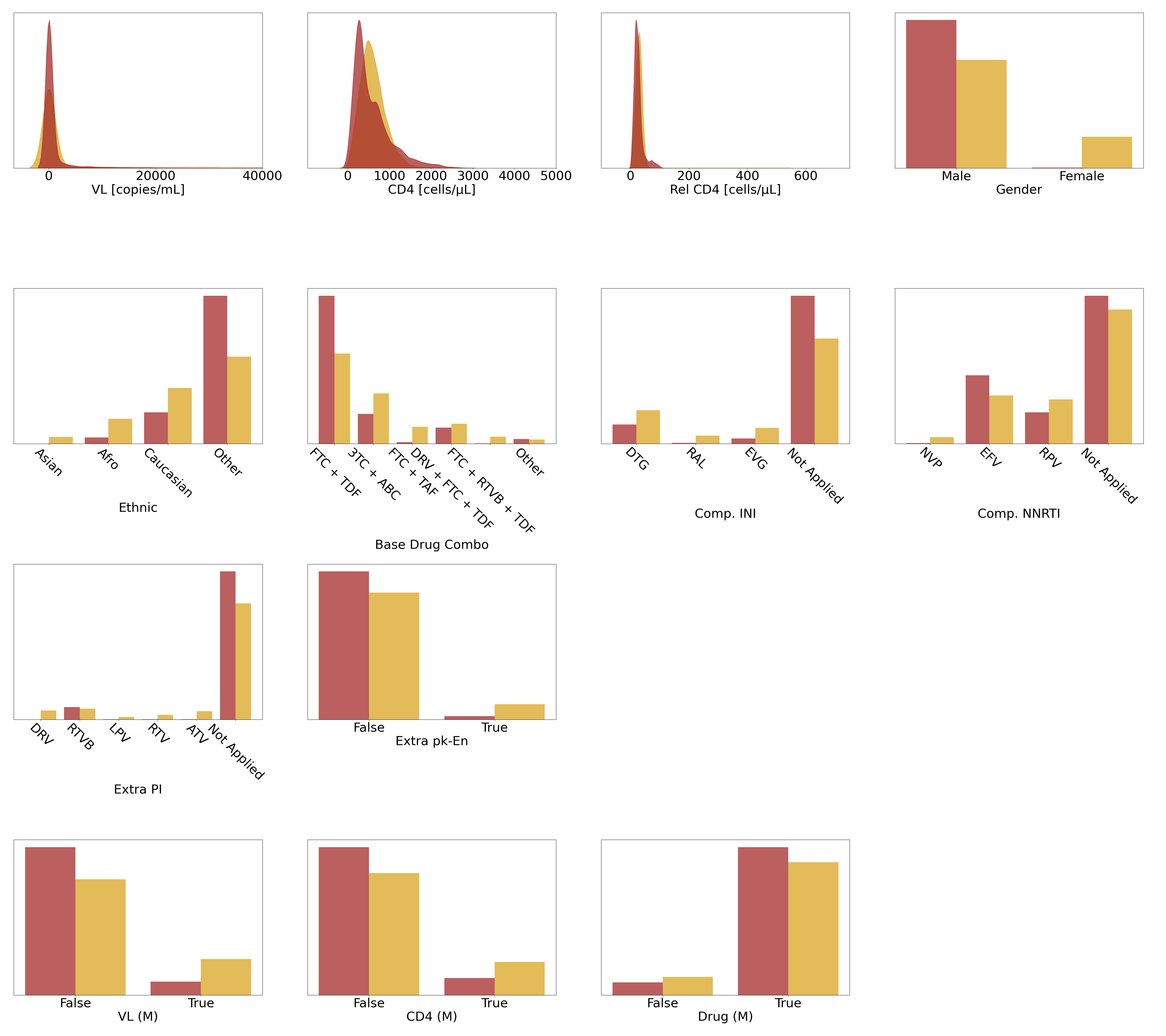}
      \caption{Synthetic dataset $\mathfrak{D}_\text{null}$ from \textcolor{brown}{\texttt{WGAN-GP}} \citep{kuo2022health} in brown.}
    \end{subfigure}
    
    \begin{subfigure}{0.825\linewidth}
      \centering
      \includegraphics[width=\linewidth]{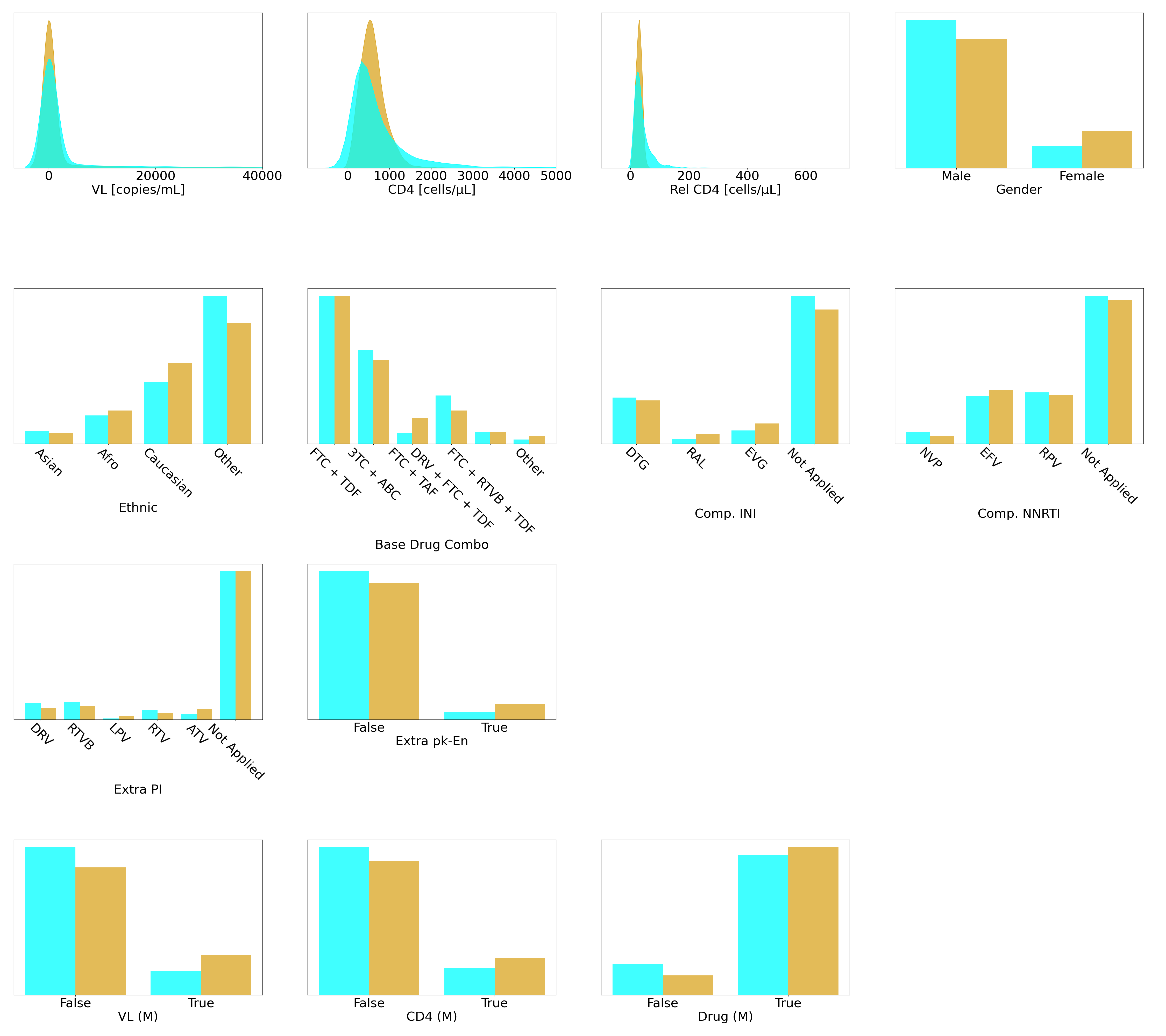}
      \caption{Synthetic dataset $\mathfrak{D}_\text{alt}$ from \textcolor{cyan}{\texttt{WGAN-GP+G\_EOT+VAE+Buffer}} (ours) in cyan.}
    \end{subfigure}%
    
    \caption{\label{Fig:KdeBar}Comparing the individual variable distributions, with the real dataset in colour gold.}
\end{figure}

Figure \ref{Fig:KdeBar} presents the KDE plots and side-by-side barplots for the individual variable comparisons. The real variables from $\mathfrak{D}_\text{real}$ are in gold; subplot \ref{Fig:KdeBar}(a) illustrates the synthetic variables in $\mathfrak{D}_\text{null}$ generated using \textcolor{brown}{\texttt{WGAN-GP}}~\cite{kuo2022health} in brown, and subplot \ref{Fig:KdeBar}(b) depicts those in $\mathfrak{D}_\text{alt}$ generated with our \textcolor{cyan}{\texttt{WGAN-GP+G\_EOT+VAE+Buffer}} in cyan. 

Overall, the distributions in subplot \ref{Fig:KdeBar}(b) are more similar than those in subplot \ref{Fig:KdeBar}(a). Thus our dataset $\mathfrak{D}_\text{alt}$ captures more details in the ground truth $\mathfrak{D}_\text{real}$ than \citet{kuo2022health}'s $\mathfrak{D}_\text{null}$. Specifically, our $\mathfrak{D}_\text{alt}$ is more capable of representing females in gender; Asians in ethnicity; and the prescription of less common medications (\textit{e.g.,} NVP in the NNRTI medication class and DRV in the PI medication class). In contrast, $\mathfrak{D}_\text{null}$ exhibits a bias towards the dominant class in the binary and categorical variables (\textit{e.g.,} making the medication in $\mathfrak{D}_\text{null}$ even more unlikely to include Extra pk-En than that in $\mathfrak{D}_\text{real}$). This hence shows that the additional VAE and extra buffer in our setup was effective in capturing extreme class imbalanced distributions in real world clinical data. Refer to Appendix \ref{Sec:HIVExtra} for extra results on our \textcolor{magenta}{\texttt{WGAN-GP+VAE+Buffer}} setup.

We then examined the synthetic variables using a series of hierarchically structured statistical tests outlined in Section \ref{Sec:RealInd}. All variables of the \textcolor{cyan}{\texttt{WGAN-GP+G\_EOT+VAE+Buffer}} synthetic dataset passed the KS test revealing that all variables of our synthetic dataset are realistic and capture both the mean and variance of their real counterparts. In contrast, the synthetic VL distribution in \citeauthor{kuo2022health}'s \textcolor{brown}{\texttt{WGAN-GP}} generated dataset failed the KS test because it was not able to mirror the spread of the real VL distribution \footnote{See Tab. 7 on page 26 of \citet{kuo2022health}.}. Refer to Appendix \ref{Sec:HIVExtra} for the complete statistical outcomes.

\subsection{Correlations among Variables} 

\begin{figure}[ht!]
    \centering
    \begin{subfigure}{\linewidth}
      \centering
      \includegraphics[width=\linewidth]{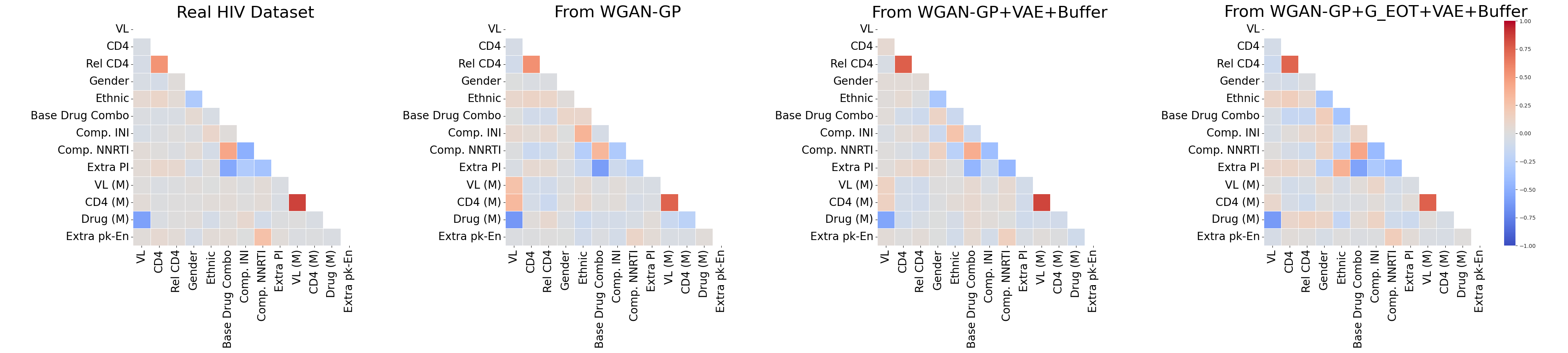}
      \caption{The classic static correlation.}
    \end{subfigure}
    
    \begin{subfigure}{\linewidth}
      \centering
      \includegraphics[width=\linewidth]{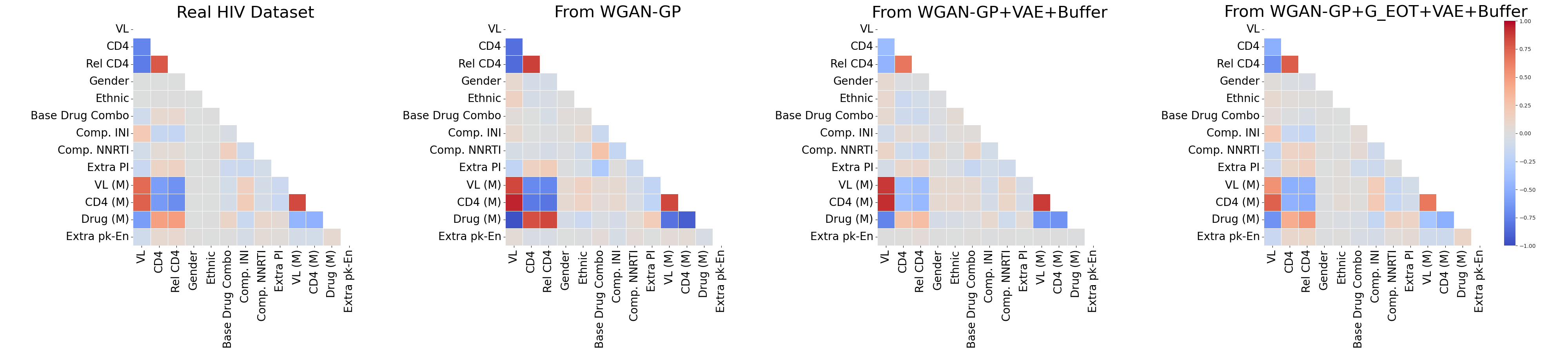}
      \caption{The dynamic correlation in trends.}
    \end{subfigure}%
    
    \begin{subfigure}{\linewidth}
      \centering
      \includegraphics[width=\linewidth]{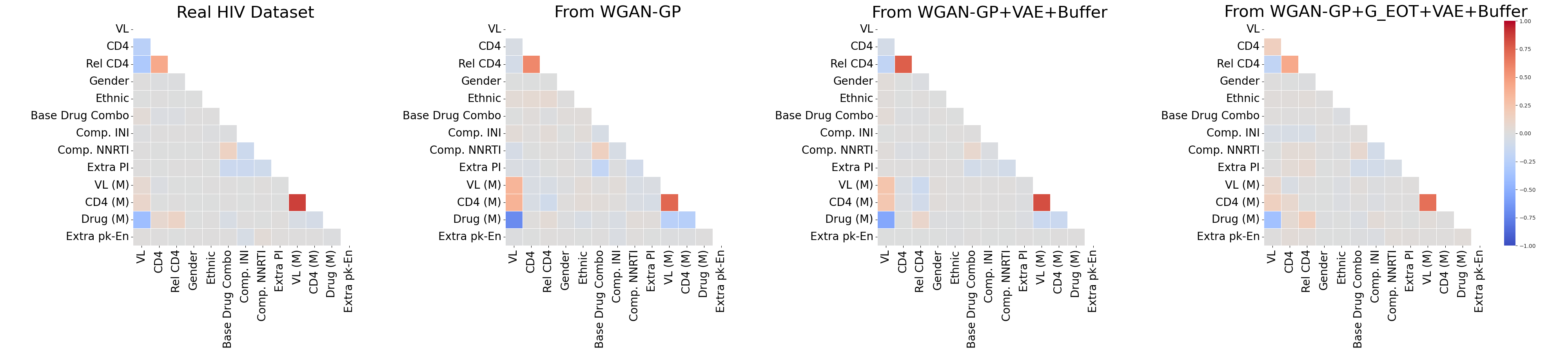}
      \caption{The dynamic correlation in cycles.}
    \end{subfigure}%
    
    \caption{\label{Fig:Correlations}Comparing different types of correlations in the real and synthetic datasets.}
\end{figure}

We continued to check the fidelity among variable pairs following the procedure described in Section \ref{Sec:FidelCorr}. The correlations among the variables of the datasets are shown in Figure \ref{Fig:Correlations}. Overall, \citet{kuo2022health}'s \textcolor{brown}{\texttt{WGAN-GP}} and our two extended setups were all able to generate synthetic datasets with realistic correlations. This applied to both the static correlation in subplot \ref{Fig:Correlations}(a) and the dynamic correlations in subplots \ref{Fig:Correlations}(b) and (c). 

While less significant, it can be argued that the variables in $\mathfrak{D}_\text{null}$ generated using the baseline \textcolor{brown}{\texttt{WGAN-GP}} have the tendency to increase the magnitudes of correlations. Some examples of this

\newpage

\newpage

behaviour can be found in the pairs of (Drug (M), CD4) and (Extra PI, Base Drug Combo) in the dynamic correlation in trends; and likewise for (CD4 (M), VL) in the dynamic correlations in cycles.  

In contrast, it could be argued that the correlations between variables in $\mathfrak{D}_\text{alt}$ generated using our \textcolor{cyan}{\texttt{WGAN-GP+G\_EOT+VAE+Buffer}} tend to be weaker than those in the real dataset. This is observed in (CD4, VL) in the dynamic correlation in trends and similarly in (CD4 (M), VL (M)) in the dynamic correlation in cycles.

\subsection{The Patient Disclosure Risk}
Since our aim is to create realistic synthetic data available for public access, we evaluated the risk of patient re-identification as discussed in Section \ref{Sec:PDR}. The minimum Euclidean distance between the real dataset and the synthetic dataset generated using \textcolor{magenta}{\texttt{WGAN-GP+VAE+Buffer}} is $0.1029$. It is $0.1229$ with the synthetic dataset generated via \textcolor{cyan}{\texttt{WGAN-GP+G\_EOT+VAE+Buffer}}. Hence no real record is leaked into the synthetic dataset using either of our setups. Note, it is not meaningful to compare the magnitudes of the Euclidean distances and what we desire is that both $>0$. 

Using \citet{el2020evaluating}'s metric, the disclosure risk of \textcolor{magenta}{\texttt{WGAN-GP+VAE+Buffer}}'s synthetic dataset is $0.081\%$ while it is $0.084\%$ for \textcolor{cyan}{\texttt{WGAN-GP+G\_EOT+VAE+Buffer}}'s synthetic dataset. In comparison, the risk is $0.041\%$ for \textcolor{brown}{\texttt{WGAN-GP}} (see page 9 of \citet{kuo2022health}). All of the results are much lower than the $9\%$ threshold suggested by the \citet{world2016consolidated}. Thus combined with our prior results in this section, we conclude that our synthetic ART for HIV datasets generated using our extended WGAN-GP setup are both realistic and secure.

\subsection{Data Utility}\label{Sec:Uti02Z}
We followed the descriptions in Section \ref{Sec:Uti01} and tested the utility of the synthetic datasets via training RL agents to suggest clinical treatments. We visualised the relative frequencies of the 

\newpage
\begin{figure}[ht!]
    \centering
    \begin{subfigure}{0.9\linewidth}
      \centering
      \includegraphics[width=\linewidth]{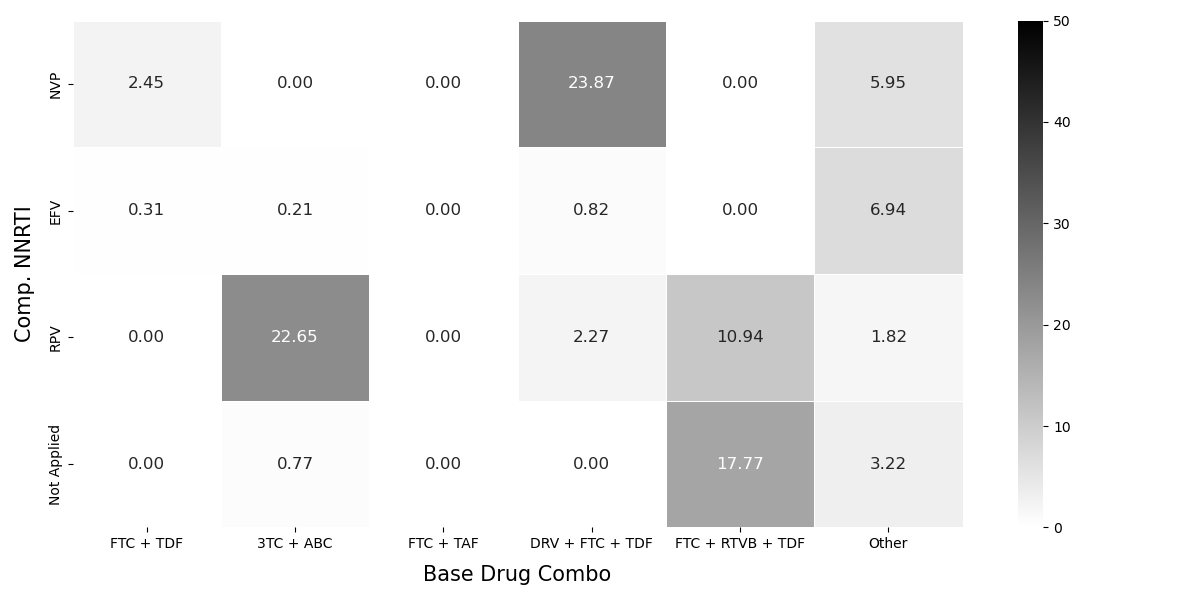}
      \caption{Trained using the real dataset $\mathfrak{D}_\text{real}$.}
    \end{subfigure}
    
    \begin{subfigure}{0.9\linewidth}
      \centering
      \includegraphics[width=\linewidth]{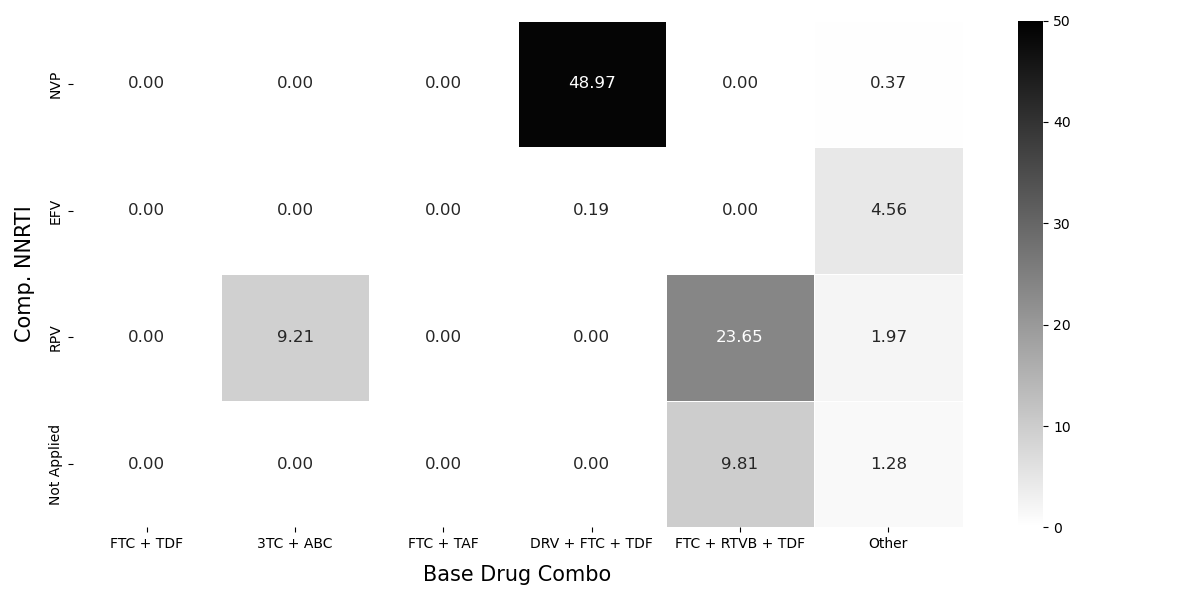}
      \caption{Trained using the synthetic dataset $\mathfrak{D}_\text{null}$ generated from \textcolor{brown}{\texttt{WGAN-GP}} \citep{kuo2022health}.}
    \end{subfigure}
    
    \begin{subfigure}{0.9\linewidth}
      \centering
      \includegraphics[width=\linewidth]{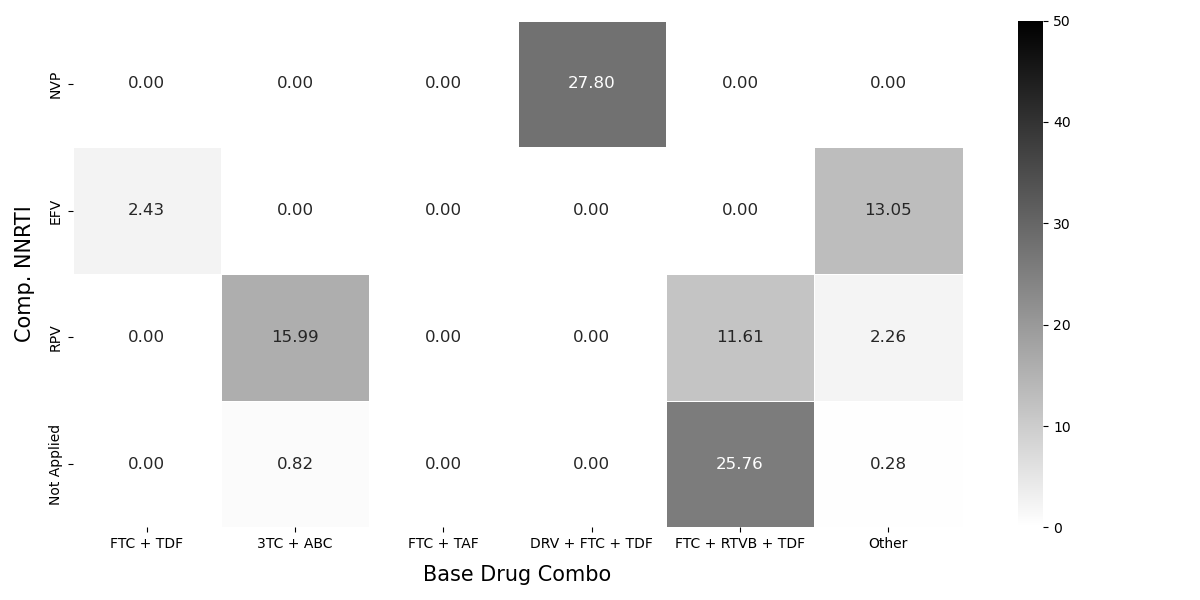}
      \caption{Trained using the synthetic dataset $\mathfrak{D}_\text{alt}$ generated from our \textcolor{cyan}{\texttt{WGAN-GP+G\_EOT+VAE+Buffer}}.}
    \end{subfigure}
    
    \caption{\label{Fig:UtilityP1}The suggestions made by RL agents trained on different ART for HIV datasets with Comp. NNRTI and Base Drug Combo spanning the action space.}
\end{figure}

\newpage
\begin{figure}[ht!]
    \centering
    \begin{subfigure}{0.9\linewidth}
      \centering
      \includegraphics[width=\linewidth]{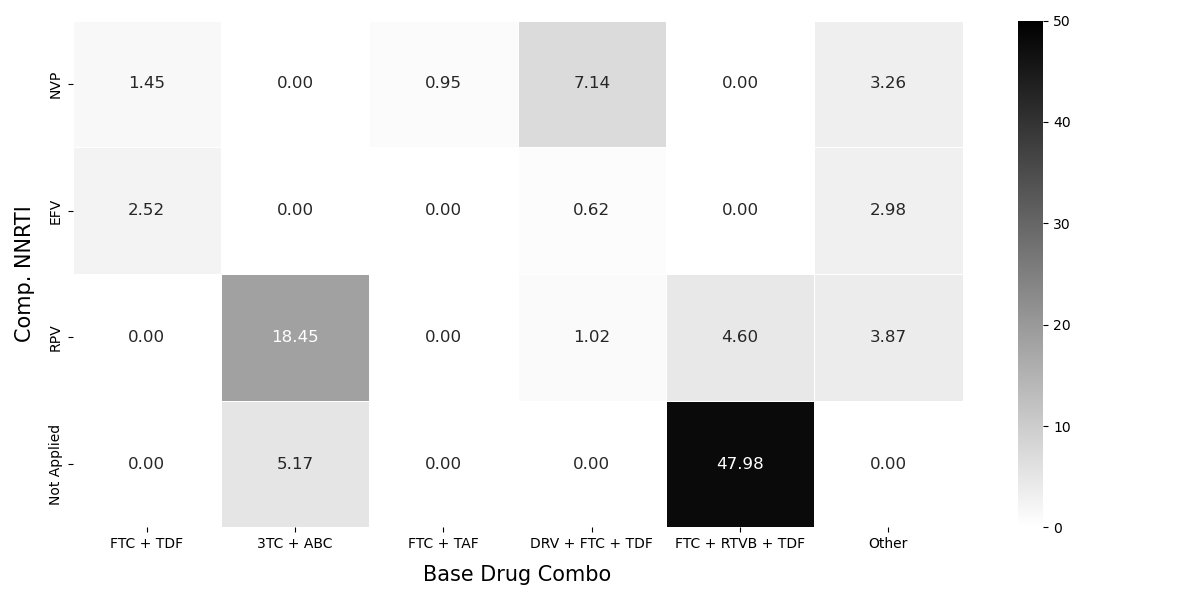}
      \caption{Trained using the real dataset $\mathfrak{D}_\text{real}$.}
    \end{subfigure}
    
    \begin{subfigure}{0.9\linewidth}
      \centering
      \includegraphics[width=\linewidth]{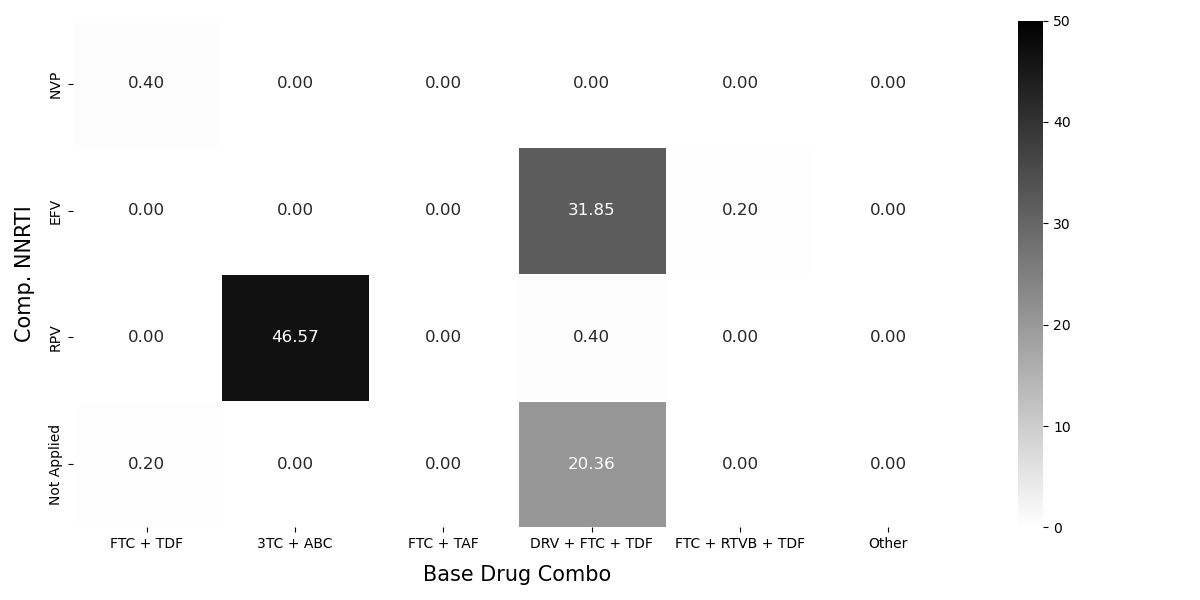}
      \caption{Trained using the synthetic dataset $\mathfrak{D}_\text{null}$ generated from \textcolor{brown}{\texttt{WGAN-GP}} \citep{kuo2022health}.}
    \end{subfigure}
    
    \begin{subfigure}{0.9\linewidth}
      \centering
      \includegraphics[width=\linewidth]{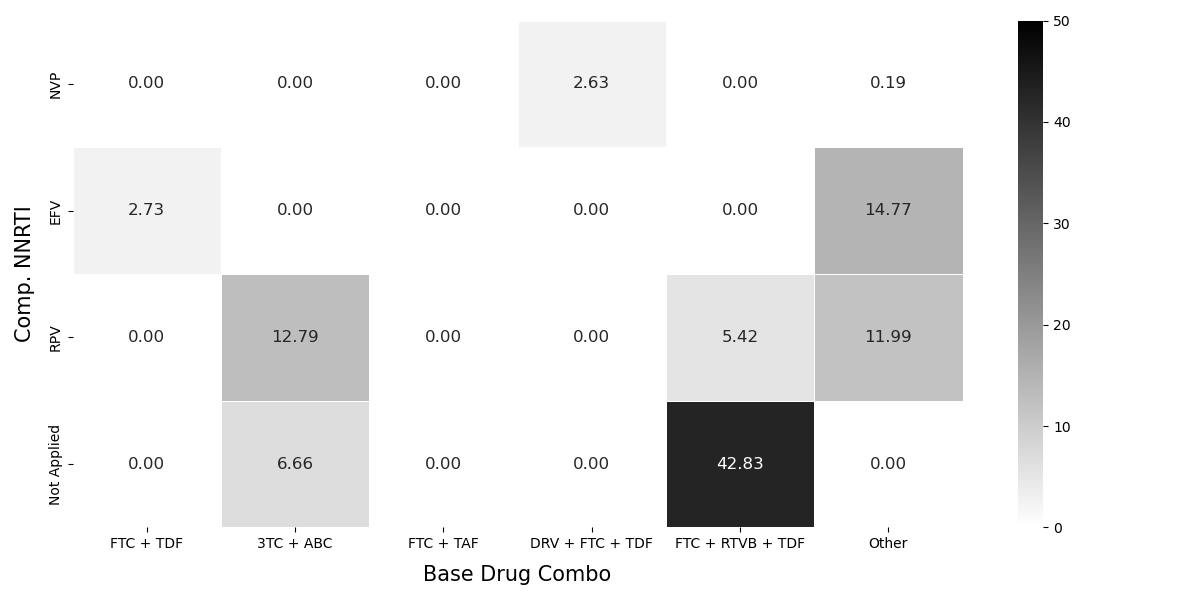}
      \caption{Trained using the synthetic dataset $\mathfrak{D}_\text{alt}$ generated from our \textcolor{cyan}{\texttt{WGAN-GP+G\_EOT+VAE+Buffer}}.}
    \end{subfigure}
    
    \caption{\label{Fig:UtilityP02}The suggestions made by RL agents trained on African patients in different ART for HIV datasets with Comp. NNRTI and Base Drug Combo spanning the action space.}
\end{figure}

\newpage
actions taken by the trained RL agents using heatmaps. Each tile represents a unique action, and the number on the tile represents the frequency of that action, as a proportion of all actions. This section primarily focuses on the synthetic dataset generated using \textcolor{cyan}{\texttt{WGAN-GP+G\_EOT+VAE+Buffer}} as our default $\mathfrak{D}_\text{alt}$.

\subsubsection{General Utility}\label{Sec:GenUt}
Figure \ref{Fig:UtilityP1} illustrates the actions taken by the RL agents when the action space was spanned by Comp. NNRTI and Base Drug Combo. Subplot (a) presents the actions taken by an RL agent trained using the real dataset $\mathfrak{D}_\text{real}$; subplot (b) is when the RL agent was trained on the synthetic dataset $\mathfrak{D}_\text{null}$ generated by \textcolor{brown}{\texttt{WGAN-GP}} \citep{kuo2022health}; and likewise subplot (c) for an RL agent trained on the synthetic dataset $\mathfrak{D}_\text{alt}$ generated using our \textcolor{cyan}{\texttt{WGAN-GP+G\_EOT+VAE+Buffer}}. 

The heatmap in subplot (b) does not match the one in subplot (a), indicating that the RL agent trained using $\mathfrak{D}_\text{null}$ generated from \textcolor{brown}{\texttt{WGAN-GP}} was incapable of suggesting similar actions to the RL agent trained using $\mathfrak{D}_\text{real}$. The RL agent trained on $\mathfrak{D}_\text{null}$ suggested NVP for Comp. NRTI with DRV + FTC + TDF for Base Drug Combo for 48.97\% of all its actions. The undiversified policy was likely induced by mode collapse in \textcolor{brown}{\texttt{WGAN-GP}} -- causing $\mathfrak{D}_\text{null}$ to capture only a fraction of all possible treatments prescribed in real life (see Table \ref{Tab:Metrics_HIV} and Figure \ref{Fig:EthGen}). 

In contrast, subplot (c) shows that the RL agent trained using $\mathfrak{D}_\text{alt}$ generated by\\
\textcolor{cyan}{\texttt{WGAN-GP+G\_EOT+VAE+Buffer}} exhibited a higher diversification in its strategy to suggest treatment. The heat map in subplot (c) mirrored the one in subplot (a) better, showing that our synthetic dataset $\mathfrak{D}_\text{alt}$ possesses a higher utility than the baseline $\mathfrak{D}_\text{null}$. See more results in Appendix \ref{App:INI}, where the action space is spanned by Comp. INI and Base Drug Combo. 

\subsubsection{Utility in Minority Groups}\label{Sec:GenMinor}

Mode collapse in GANs has a particularly negative impact on the downstream model utility for minority groups. To demonstrate the severity, we repeated the experiments in Section \ref{Sec:GenUt} but only with patients of African ethnicity. The results for the experiments are shown in Figure \ref{Fig:UtilityP02}.

As previously demonstrated in Figure \ref{Fig:EthGen}, \citet{kuo2022health}'s \textcolor{brown}{\texttt{WGAN-GP}} experienced mode collapse and had difficulties in generating synthetic patients of Asian and African ethnicity. This meant that there weren't many data points in $\mathfrak{D}_\text{null}$ to cover the diversity in the ART for HIV regimens for these minority patients. As a result, subplot \ref{Fig:UtilityP02}(b) differs greatly from subplot \ref{Fig:UtilityP02}(a) and the RL agent trained using $\mathfrak{D}_\text{null}$ was incapable of suggesting similar actions to the RL agent trained using $\mathfrak{D}_\text{real}$. 

In stark contrast, subplot \ref{Fig:UtilityP02}(c) captures most features that can be found in subplot \ref{Fig:UtilityP02}(a). This shows that our synthetic dataset $\mathfrak{D}_\text{alt}$ has high utility and is more suitable to replace the baseline $\mathfrak{D}_\text{null}$ for supporting the development of downstream machine learning algorithms. As discussed in Section \ref{Sec:MCTS}, this is made possible because our \textcolor{cyan}{\texttt{WGAN-GP+G\_EOT+VAE+Buffer}} was effective in mitigating mode collapse in generating real world clinical data.

\section{Discussion}
This paper introduced techniques to support WGAN-GP in mitigating mode collapse when synthesising realistic clinical time-series datasets. Inspired by the work of \citet{larsen2016autoencoding} (see Sections \ref{Sec:OurShit} and \ref{Sec:ReplaySample}), we introduced an additional VAE to the conventional WGAN-GP setup to forward information about the latent structure of the real data to the GAN generator. Furthermore, in order to enhance GAN's ability to generate realistic data, we decided to employ an extra buffer (see Section \ref{Sec:OurShit}) to replay the exact combinations of latent features of real data (in an otherwise large and sparse feature representation space).

We tested seven variants of GAN setups (see Sections \ref{Sec:Baselines} and \ref{Sec:ResultsHIV}) -- comparing both our techniques against \citet{kuo2022health}'s \textcolor{brown}{\texttt{WGAN-GP}} and prior methods that mitigated mode collapse for image generation. We found that all baselines were not effective in mitigating mode collapse for clinical time series data; hence they could not capture details of all imbalanced variables (see Figure \ref{Fig:EthGen} and Table \ref{Tab:Metrics_HIV}). Using ART for HIV from the EuResist database \citep{zazzi2012predicting} as a case study, we found that mode collapse in the baseline GANs caused the models to neglect the creation of synthetic patients of minority demographics (Asians and Africans, see Figure \ref{Fig:EthGen}(b)). Likewise, we also observed that mode collapse caused the baseline GANs to not be able to learn the imbalanced distribution in medication usage (see Figure \ref{Fig:KdeBar}(a)). In stark contrast, our synthetic datasets generated using our extended GAN setup (\textit{i.e.,} \textcolor{cyan}{\texttt{WGAN-GP+G\_EOT+VAE+Buffer}}) achieved a higher level of synthetic patient cohort diversity.

We further showed that the diversity in synthetic datasets greatly impacted their utility for supporting the development of downstream model building (see Section \ref{Sec:Uti02Z}). RL agents trained using our synthetic dataset $\mathfrak{D}_\text{alt}$ were able to suggest similar actions to those trained on the real dataset $\mathfrak{D}_\text{real}$. This was not achieved using \citet{kuo2022health}'s synthetic dataset $\mathfrak{D}_\text{null}$ as their \textcolor{brown}{\texttt{WGAN-GP}} experienced mode collapse. The differences in the suggested actions were particularly different when treating patients of minority background (\textit{e.g.,} African, see Figure \ref{Fig:UtilityP02}); showing that synthetic datasets generated from a GAN model that experienced mode collapse could potentially cause indirect harm in patient care.

The proposed approach improves both the \textit{resemblance} and the \textit{representation}~\citep{bhanot2021problem} of synthetic data generated by GAN-based methods. Resemblance indicates how closely the synthetic data match the real data, and representation is used in the context of fairness to indicate how adequately priority populations are represented in the generated data~\citep{thomasian2021advancing}. These concepts are receiving increasing attention and echo the ethic guidelines in artificial intelligence published by the \citet{AussieAIEthics} and the \citet{USAIEthics}.

The technique introduced in this paper is currently only tested on ART for HIV. In future work, we aim to test our \textcolor{cyan}{\texttt{WGAN-GP+G\_EOT+VAE+Buffer}} model on additional clinical datasets. This could be particularly interesting because \citet{kuo2022health} noted that while their \textcolor{brown}{\texttt{WGAN-GP}} had some shortcomings in generating ART for HIV data, \textcolor{brown}{\texttt{WGAN-GP}} performed favourably on acute hypotension \citep{gottesman2020interpretable} and sepsis \citep{komorowski2018artificial} datasets. In contrast to ART for HIV, the latter datasets mainly comprise numeric variables. The appropriate use cases of \textcolor{cyan}{\texttt{WGAN-GP+G\_EOT+VAE+Buffer}} should thus be further analysed.

While we found our synthetic dataset to be realistic, safe, and to have high utility, the generated synthetic dataset should still not be regarded as a direct replacement for the real dataset. In this paper, we used batch-constrained Q-learning to evaluate whether our synthetic dataset could be used to train RL agents; and in future work, we intend to conduct a more thorough evaluation covering more complicated algorithms including recent advancements in offline RL. Also in future work, we aim to compare GAN-based models with alternative generative models including diffusion models \citep{dhariwal2021diffusion}.

\section{Conclusion}
This paper introduced an additional VAE and an extra buffer to support WGAN-GP in synthesising realistic clinical time-series datasets. Using ART for HIV as a case study, we demonstrated that our resulting synthetic dataset
\begin{itemize}[noitemsep,topsep=0pt]
    \item achieves a better level of cohort diversity;
    \item mirrors the distributions of all real variables;
    \item accurately captures the correlations among the real variables;
    \item has high security and protects the identity of real patients; and
    \item possesses high utility to support the development of downstream machine learning models.
\end{itemize}
Of note, improvements in synthetic data quality were particularly marked in data subsets related to minority groups.

\textbf{Data Access and Descriptions}\\
Our synthetic dataset generated using \textcolor{cyan}{\texttt{WGAN-GP+G\_EOT+VAE+Buffer}} is hosted on our website \url{https://healthgym.ai/} and is publicly accessible. The synthetic ART for HIV dataset is 44.7 MB; and is stored as a \textit{comma separated value} (CSV) file. 

There are 8,916 patients in the data and records for each patient span 60 months. The data is summarised monthly; there are hence 534,960 (=8,916 $\times$ 60) rows (records) in total. There are 15 columns -- including the 13 ART for HIV variables as shown in Table \ref{Tab:VarsOfHIV}, a variable indicating the patient identifier, and a variable specifying the time point.

\newpage
\typeout{}
\bibliography{iclr2021_conference}
\bibliographystyle{iclr2021_conference}

\section*{Author Contributions Statement}
\textbf{N.K.} and \textbf{S.B.} designed, implemented and validated the deep learning models used to generate the synthetic datasets. \textbf{L.J.} contributed to the design of the study and provided expertise regarding the risk of sensitive information disclosure. \textbf{M.P.} provided clinical expertise on antiretroviral therapy for HIV. \textbf{F.G.}, \textbf{A.S.}, \textbf{M.Z.}, \textbf{M.B.}, and \textbf{R.K.} contributed patient data as part of the EuResist Integrated Database. Furthermore, \textbf{N.K.} wrote the manuscript and \textbf{S.B.} and \textbf{N.K.} designed the study. All authors contributed to the interpretation of findings and manuscript revisions.

\section*{Competing Interests}
The authors declare no competing interests.

\section*{Acknowledgements}
This study benefited from data provided by EuResist Network EIDB; and this project has been funded by a Wellcome Trust Open Research Fund (reference number 219691/Z/19/Z).

\newpage
\appendix
\section*{Supplementary Materials}

\textbf{Nicholas I-Hsien Kuo$^{1}$},\\
\textbf{Federico Garcia}$^{2}$, \textbf{Anders S\"onnerborg}$^{3}$, \textbf{Maurizio Zazzi}$^{4}$, \textbf{Michael B\"ohm}$^{5}$, \textbf{Rolf Kaiser}$^{5}$,\\
\textbf{Mark Polizzotto}$^{6}$, \textbf{Louisa Jorm}$^{1}$, \textbf{Sebastiano Barbieri}$^{1}$\\
$^{1}$Centre for Big Data Research in Health (CBDRH), the University of New South Wales,\\
\hspace*{2mm}Sydney, Australia\\
$^{2}$Hospital Universitario San Cecilio, Granada, Spain\\
$^{3}$Hospital Karolinska Institutet, Stockholm, Sweden\\
$^{4}$Universit{\`a} degli Studi di Siena, Siena, Italy\\
$^{5}$Uniklinik K{\"o}ln, Universit{\"a}t zu K{\"o}ln, Cologne, Germany\\
$^{6}$Australian National University, Canberra, Australia\\
\textcolor{white}{*}\\
Corresponding author: Nicholas I-Hsien Kuo (\texttt{n.kuo@unsw.edu.au})\\

In this study, we presented an extended setup to GAN models~\citep{goodfellow2014generative} for synthesising clinical time series data. We based our work on \citet{kuo2022health}'s WGAN-GP and demonstrated the effectiveness in mitigating mode collapse~\citep{goodfellow2016nips} on the antiretroviral therapy for human immunodeficiency virus (ART for HIV)~\citep{zazzi2012predicting}. We conducted several experiments with different setups; and the supplementary materials include some additional interesting secondary results.

\section{On the Realisticness of Individual Variables}\label{Sec:HIVExtra}
\textbf{Extra Results on \textcolor{magenta}{\texttt{WGAN-GP+VAE+Buffer}}}\\
\begin{figure}[ht!]
    \centering

    \includegraphics[width=0.825\linewidth]{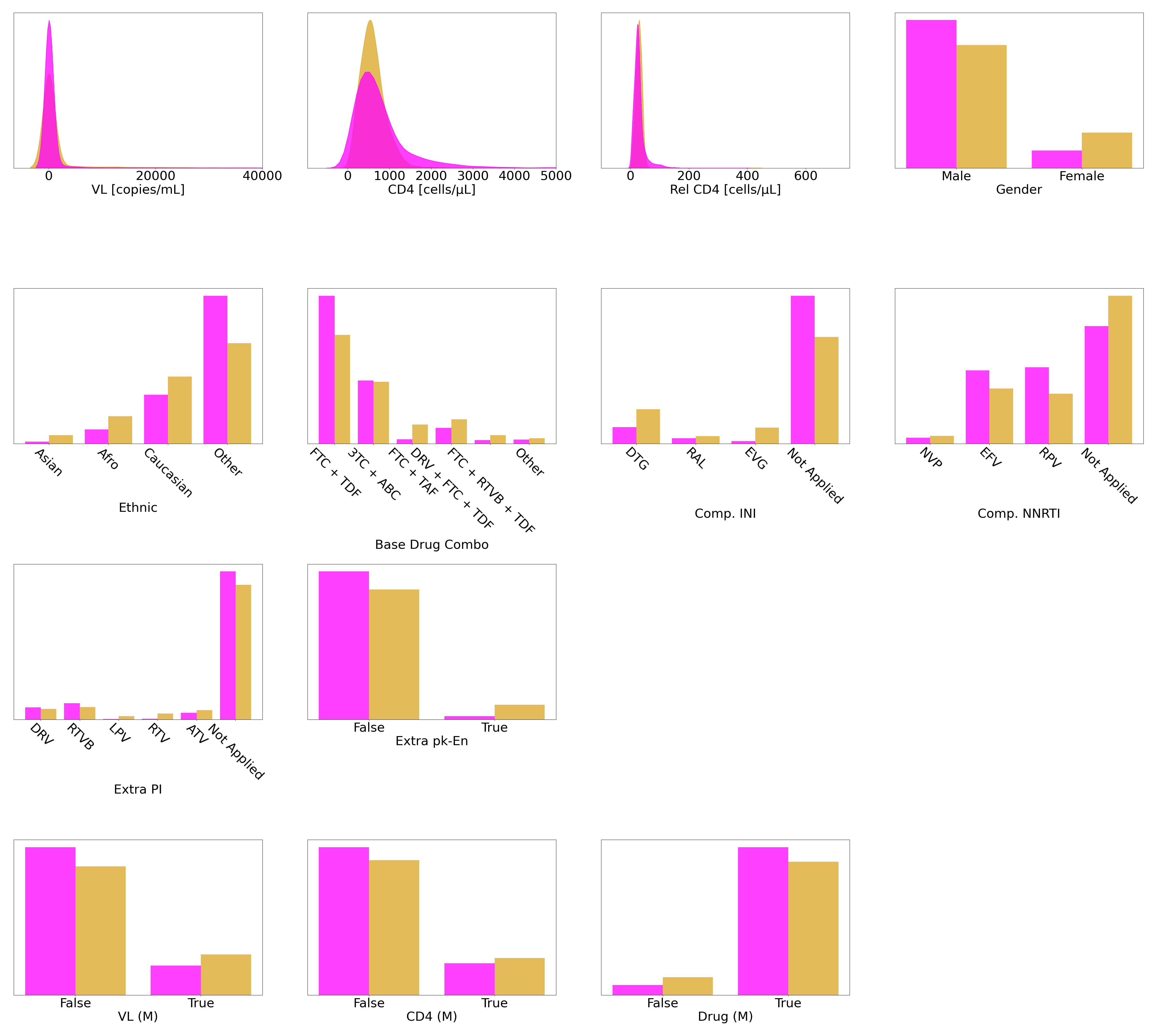}
    
    \caption{\label{Fig:KdenBarAnother}Comparing the individual variables of the real dataset (in colour gold) with those in the synthetic dataset generated with \textcolor{magenta}{\texttt{WGAN-GP+VAE+Buffer}} (in magenta).}
\end{figure}

Following Section \ref{Sec:RIV}, we plotted the variable distributions of the synthetic dataset generated using our \textcolor{magenta}{\texttt{WGAN-GP+VAE+Buffer}} in Figure \ref{Fig:KdenBarAnother}. Compared to the baseline \textcolor{brown}{\texttt{WGAN-GP}}~\citep{kuo2022health} in Figure \ref{Fig:KdeBar}(a), we observed that \textcolor{magenta}{\texttt{WGAN-GP+VAE+Buffer}} was more capable of capturing severe class imbalance. However, the results were less good than our \textcolor{cyan}{\texttt{WGAN-GP+G\_EOT+VAE+Buffer}} setup shown in Figure \ref{Fig:KdeBar}(b); this model experienced some notable difficulties in representing categorical variables with many levels such as Extra PI and Comp. INI.

\textbf{The Complete Statistical Outcome}\\
We followed \citet{kuo2022health}'s work and computed a series of hierarchical statistical tests (see Figure \ref{Fig:HST}) to verify the realisticness of the synthetic variables. The results are organised in Tables \ref{Tab:HstHivHealthGym}, \ref{Tab:HstHivOurs1}, and \ref{Tab:HstHivOurs2}. The purpose of the statistical tests was to check whether an arbitrary neural network would consider the synthetic dataset highly similar to the real dataset during iterative batch training. Hence for a maximum of $100$ iterations (corresponding to $100$ in the denominator in the tables), we sampled a batch of synthetic and real data each of batch size $32$, and performed at most 4 tests along the variable dimension. Refer to App. D.5 on page 44 of \citet{kuo2022health} for the full algorithmic description.

The biggest problem in \citet{kuo2022health}'s synthetic dataset was VL which had a misaligned variance. This concern is now addressed in our datasets. While all variables in the synthetic dataset generated by our \textcolor{magenta}{\texttt{WGAN-GP+VAE+Buffer}} are realistic, we observed that there are still some misalignments in variances. Notably, the synthetic Ethnicity variable passed less than $70\%$ of F-test (Table \ref{Tab:HstHivOurs1}), hinting that there was a slight mode collapse problem which may have caused mismatches among the categories for this variable. In contrast, no major problem was observed in Table \ref{Tab:HstHivOurs2}; hence all variables generated using our better performing model of \textcolor{cyan}{\texttt{WGAN-GP+G\_EOT+VAE+Buffer}} are realistic.

\section{Extra Results on Data Utility}\label{App:INI}

\begin{figure}[ht!]
    \centering
    \begin{subfigure}{0.9\linewidth}
      \centering
      \includegraphics[width=\linewidth]{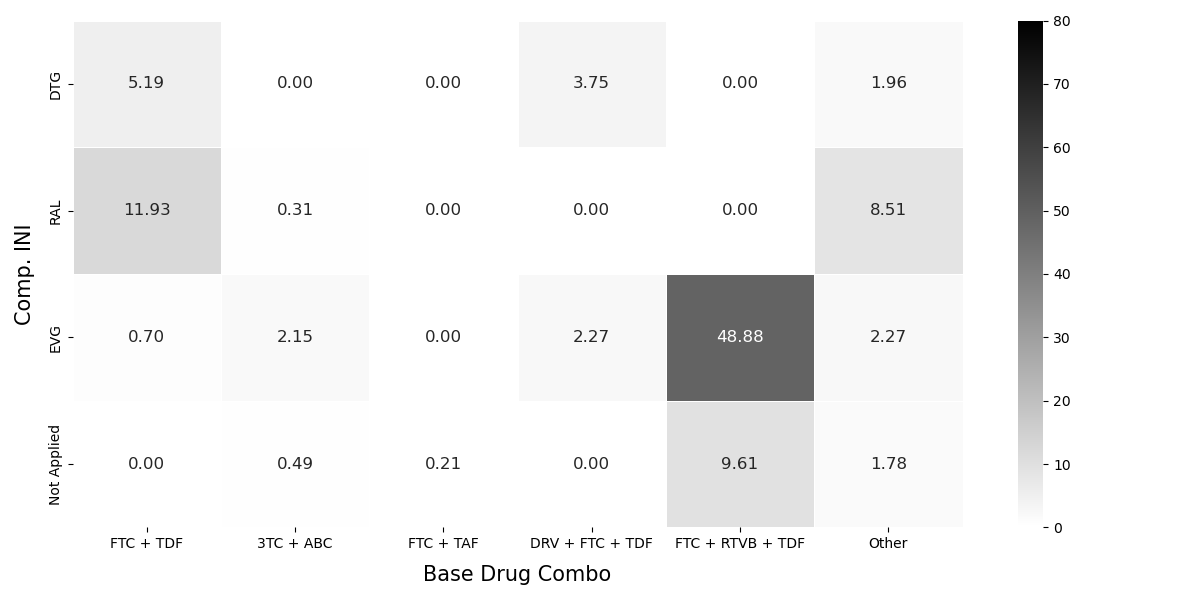}
      \caption{Trained using the real dataset $\mathfrak{D}_\text{real}$.}
    \end{subfigure}
    
    \begin{subfigure}{0.9\linewidth}
      \centering
      \includegraphics[width=\linewidth]{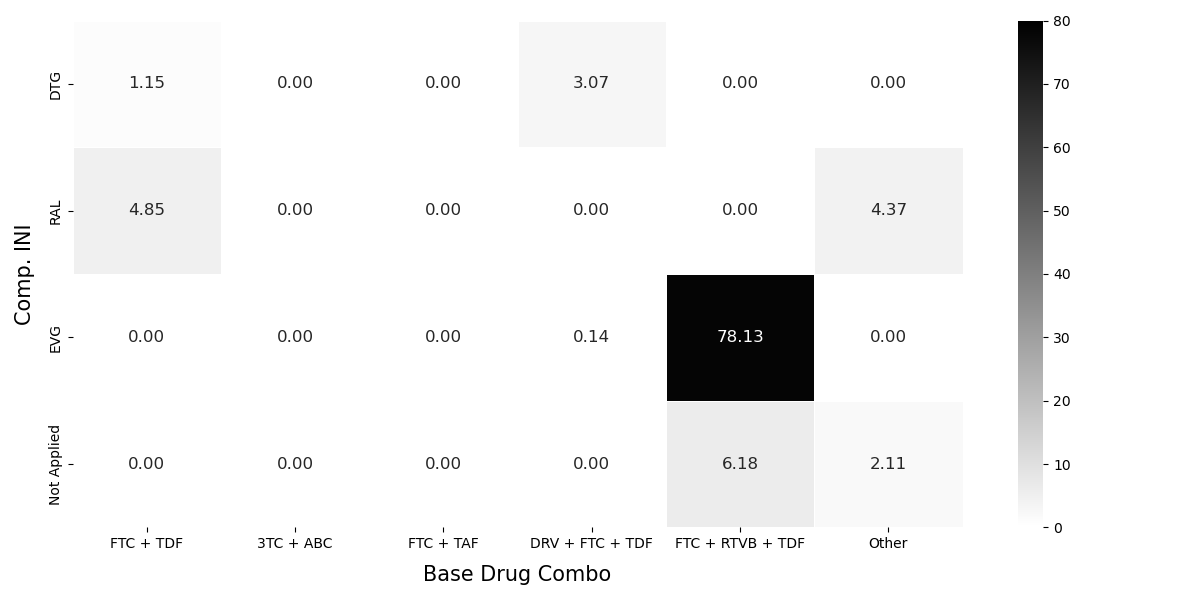}
      \caption{Trained using the synthetic dataset $\mathfrak{D}_\text{null}$ generated from \textcolor{brown}{\texttt{WGAN-GP}} \citep{kuo2022health}.}
    \end{subfigure}
    
    \begin{subfigure}{0.9\linewidth}
      \centering
      \includegraphics[width=\linewidth]{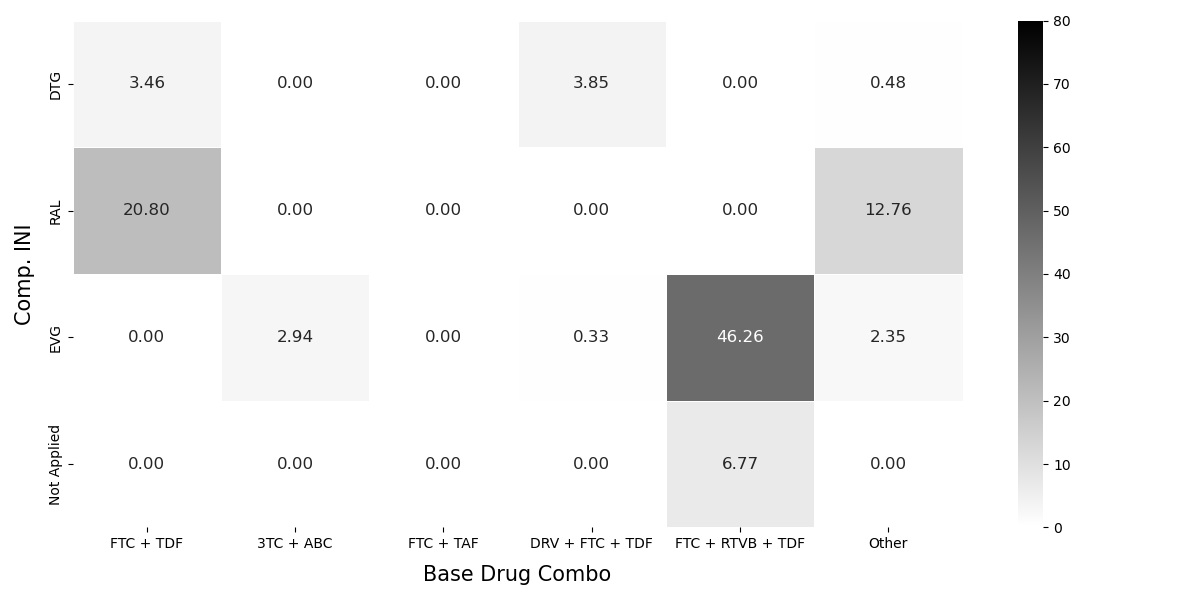}
      \caption{Trained using the synthetic dataset $\mathfrak{D}_\text{alt}$ generated from our \textcolor{cyan}{\texttt{WGAN-GP+G\_EOT+VAE+Buffer}}.}
    \end{subfigure}
    
    \caption{\label{Fig:UtilityApp}The suggestions made by RL agents trained on different ART for HIV datasets with Comp. INI and Base Drug Combo spanning the action space.}
\end{figure}

Following Section \ref{Sec:GenUt}, we continued to train RL agents using different synthetic datasets and spanned the action space through the variables Comp. INI and Base Drug Combo. The results are shown in Figure \ref{Fig:UtilityApp}. 

Again, we see that our synthetic dataset $\mathfrak{D}_\text{alt}$ posses a higher utility than the baseline $\mathfrak{D}_\text{null}$. The RL agent trained using $\mathfrak{D}_\text{null}$ was incapable of suggesting similar treatments to the RL agent trained using $\mathfrak{D}_\text{real}$. Notably, the former would suggest up to 78.13\% of all its actions on EVG for Comp. INI and FTC + RTVB + TDF for Base Drug Combo. In contrast, the RL agent trained using our $\mathfrak{D}_\text{alt}$ suggested a more diverse treatment strategy and mirrored the behaviour of the RL agent trained using the ground truth.

\newpage
\begin{table}[ht]
    \footnotesize
    \centering
    \begin{tabular}{|l||c|c|c|c|}
        \hline
        \textbf{Variable Name} & \textbf{\textcolor{white}{.}\hspace{3mm} KS-Test \textcolor{white}{.}\hspace{3mm}} & 
        \textbf{\textcolor{white}{.}\hspace{3mm} t-Test \textcolor{white}{.}\hspace{3mm}} & 
        \textbf{\textcolor{white}{.}\hspace{3mm} F-Test \textcolor{white}{.}\hspace{3mm}} & 
        \textbf{Three Sigma Rule Test}\\
        \hline
        \hline
        VL & 
        \cellcolor{red!10}$48/100$ & 
        \cellcolor{blue!10}$90/100$ & 
        \cellcolor{red!10}$15/100$ & 
        \cellcolor{blue!10}$100/100$\\
        \hline
        CD4 & 
        \cellcolor{blue!10}$83/100$ & 
        \cellcolor{blue!10}$97/100$ & 
        \cellcolor{blue!10}$91/100$ & 
        \cellcolor{blue!10}$91/100$\\
        \hline
        Rel CD4 & \cellcolor{blue!10}$85/100$ & 
        \cellcolor{blue!10}$92/100$ & 
        \cellcolor{blue!10}$100/100$ & 
        \cellcolor{blue!10}$99/100$\\
        \hline
        \hline
        Gender & 
        \cellcolor{blue!10}$97/100$ & 
        \cellcolor{black!10}- - & 
        \cellcolor{red!10}$49/100$ &
        \cellcolor{black!10}- -\\
        \hline
        Ethnic & 
        \cellcolor{blue!10}$81/100$ & 
        \cellcolor{black!10}- - & 
        \cellcolor{red!10}$38/100$ &
        \cellcolor{black!10}- -\\
        \hline
        \hline
        Base Drug Combo & \cellcolor{blue!10}$73/100$ & 
        \cellcolor{black!10}- - & 
        \cellcolor{blue!10}$71/100$ &
        \cellcolor{black!10}- -\\
        \hline
        Comp. INI & \cellcolor{blue!10}$96/100$ & 
        \cellcolor{black!10}- - & 
        \cellcolor{blue!10}$73/100$ &
        \cellcolor{black!10}- -\\
        \hline
        Comp. NNRTI & \cellcolor{blue!10}$92/100$ & 
        \cellcolor{black!10}- - & 
        \cellcolor{blue!10}$95/100$ &
        \cellcolor{black!10}- -\\
        \hline
        Extra PI & \cellcolor{blue!10}$100/100$ & 
        \cellcolor{black!10}- - & 
        \cellcolor{blue!10}$73/100$ & 
        \cellcolor{black!10}- -\\
        \hline
        Extra pk-En & \cellcolor{blue!10}$100/100$ & 
        \cellcolor{black!10}- - & 
        \cellcolor{blue!10}$95/100$ & 
        \cellcolor{black!10}- -\\
        \hline
        \hline
        VL (M) & 
        \cellcolor{blue!10}$100/100$ & 
        \cellcolor{black!10}- - & 
        \cellcolor{blue!10}$72/100$ & 
        \cellcolor{black!10}- -\\
        \hline
        CD4 (M) & \cellcolor{blue!10}$99/100$ & 
        \cellcolor{black!10}- - & 
        \cellcolor{blue!10}$94/100$ & 
        \cellcolor{black!10}- -\\
        \hline
        Drug (M) & \cellcolor{blue!10}$100/100$ & 
        \cellcolor{black!10}- - & 
        \cellcolor{blue!10}$88/100$ & 
        \cellcolor{black!10}- -\\
        \hline
        
    \end{tabular}
    
    \caption{\label{Tab:HstHivHealthGym}The reference statistics taken from Tab. 11 on page 47 of \citet{kuo2022health}'s \textcolor{brown}{\texttt{WGAN-GP}}.}
\end{table}

\begin{table}[ht]
    \footnotesize
    \centering
    \begin{tabular}{|l||c|c|c|c|}
        \hline
        \textbf{Variable Name} & \textbf{\textcolor{white}{.}\hspace{3mm} KS-Test \textcolor{white}{.}\hspace{3mm}} & 
        \textbf{\textcolor{white}{.}\hspace{3mm} t-Test \textcolor{white}{.}\hspace{3mm}} & 
        \textbf{\textcolor{white}{.}\hspace{3mm} F-Test \textcolor{white}{.}\hspace{3mm}} & 
        \textbf{Three Sigma Rule Test}\\
        \hline
        \hline
        VL & 
        \cellcolor{blue!10}$73/100$ & 
        \cellcolor{blue!10}$95/100$ & 
        \cellcolor{red!10}$39/100$ & 
        \cellcolor{blue!10}$100/100$\\
        \hline
        CD4 & 
        \cellcolor{blue!10}$87/100$ & 
        \cellcolor{blue!10}$92/100$ & 
        \cellcolor{blue!10}$91/100$ & 
        \cellcolor{blue!10}$91/100$\\
        \hline
        Rel CD4 & \cellcolor{blue!10}$90/100$ & 
        \cellcolor{blue!10}$96/100$ & 
        \cellcolor{blue!10}$99/100$ & 
        \cellcolor{blue!10}$97/100$\\
        \hline
        \hline
        Gender & 
        \cellcolor{blue!10}$100/100$ & 
        \cellcolor{black!10}- - & 
        \cellcolor{blue!10}$74/100$ &
        \cellcolor{black!10}- -\\
        \hline
        Ethnic & 
        \cellcolor{blue!10}$89/100$ & 
        \cellcolor{black!10}- - & 
        \cellcolor{red!10}$61/100$ &
        \cellcolor{black!10}- -\\
        \hline
        \hline
        Base Drug Combo & \cellcolor{blue!10}$95/100$ & 
        \cellcolor{black!10}- - & 
        \cellcolor{blue!10}$82/100$ &
        \cellcolor{black!10}- -\\
        \hline
        Comp. INI & \cellcolor{blue!10}$97/100$ & 
        \cellcolor{black!10}- - & 
        \cellcolor{blue!10}$85/100$ &
        \cellcolor{black!10}- -\\
        \hline
        Comp. NNRTI & \cellcolor{blue!10}$98/100$ & 
        \cellcolor{black!10}- - & 
        \cellcolor{blue!10}$96/100$ &
        \cellcolor{black!10}- -\\
        \hline
        Extra PI & \cellcolor{blue!10}$100/100$ & 
        \cellcolor{black!10}- - & 
        \cellcolor{blue!10}$96/100$ & 
        \cellcolor{black!10}- -\\
        \hline
        Extra pk-En & \cellcolor{blue!10}$100/100$ & 
        \cellcolor{black!10}- - & 
        \cellcolor{red!10}$68/100$ & 
        \cellcolor{black!10}- -\\
        \hline
        \hline
        VL (M) & 
        \cellcolor{blue!10}$99/100$ & 
        \cellcolor{black!10}- - & 
        \cellcolor{blue!10}$91/100$ & 
        \cellcolor{black!10}- -\\
        \hline
        CD4 (M) & \cellcolor{blue!10}$100/100$ & 
        \cellcolor{black!10}- - & 
        \cellcolor{blue!10}$96/100$ & 
        \cellcolor{black!10}- -\\
        \hline
        Drug (M) & \cellcolor{blue!10}$100/100$ & 
        \cellcolor{black!10}- - & 
        \cellcolor{blue!10}$88/100$ & 
        \cellcolor{black!10}- -\\
        \hline
        
    \end{tabular}
    
    \caption{\label{Tab:HstHivOurs1}The statistical outcomes of our \textcolor{magenta}{\texttt{WGAN-GP+VAE+Buffer}} synthetic dataset.}
\end{table}

\begin{table}[ht]
    \footnotesize
    \centering
    \begin{tabular}{|l||c|c|c|c|}
        \hline
        \textbf{Variable Name} & \textbf{\textcolor{white}{.}\hspace{3mm} KS-Test \textcolor{white}{.}\hspace{3mm}} & 
        \textbf{\textcolor{white}{.}\hspace{3mm} t-Test \textcolor{white}{.}\hspace{3mm}} & 
        \textbf{\textcolor{white}{.}\hspace{3mm} F-Test \textcolor{white}{.}\hspace{3mm}} & 
        \textbf{Three Sigma Rule Test}\\
        \hline
        \hline
        VL & 
        \cellcolor{blue!10}$78/100$ & 
        \cellcolor{blue!10}$93/100$ & 
        \cellcolor{blue!10}$77/100$ & 
        \cellcolor{blue!10}$98/100$\\
        \hline
        CD4 & 
        \cellcolor{blue!10}$82/100$ & 
        \cellcolor{blue!10}$96/100$ & 
        \cellcolor{blue!10}$96/100$ & 
        \cellcolor{blue!10}$82/100$\\
        \hline
        Rel CD4 & \cellcolor{blue!10}$81/100$ & 
        \cellcolor{blue!10}$72/100$ & 
        \cellcolor{blue!10}$100/100$ & 
        \cellcolor{blue!10}$91/100$\\
        \hline
        \hline
        Gender & 
        \cellcolor{blue!10}$100/100$ & 
        \cellcolor{black!10}- - & 
        \cellcolor{blue!10}$87/100$ &
        \cellcolor{black!10}- -\\
        \hline
        Ethnic & 
        \cellcolor{blue!10}$90/100$ & 
        \cellcolor{black!10}- - & 
        \cellcolor{blue!10}$84/100$ &
        \cellcolor{black!10}- -\\
        \hline
        \hline
        Base Drug Combo & \cellcolor{blue!10}$95/100$ & 
        \cellcolor{black!10}- - & 
        \cellcolor{blue!10}$82/100$ &
        \cellcolor{black!10}- -\\
        \hline
        Comp. INI & \cellcolor{blue!10}$99/100$ & 
        \cellcolor{black!10}- - & 
        \cellcolor{blue!10}$92/100$ &
        \cellcolor{black!10}- -\\
        \hline
        Comp. NNRTI & \cellcolor{blue!10}$99/100$ & 
        \cellcolor{black!10}- - & 
        \cellcolor{blue!10}$95/100$ &
        \cellcolor{black!10}- -\\
        \hline
        Extra PI & \cellcolor{blue!10}$100/100$ & 
        \cellcolor{black!10}- - & 
        \cellcolor{blue!10}$95/100$ & 
        \cellcolor{black!10}- -\\
        \hline
        Extra pk-En & \cellcolor{blue!10}$100/100$ & 
        \cellcolor{black!10}- - & 
        \cellcolor{blue!10}$91/100$ & 
        \cellcolor{black!10}- -\\
        \hline
        \hline
        VL (M) & 
        \cellcolor{blue!10}$98/100$ & 
        \cellcolor{black!10}- - & 
        \cellcolor{blue!10}$71/100$ & 
        \cellcolor{black!10}- -\\
        \hline
        CD4 (M) & \cellcolor{blue!10}$100/100$ & 
        \cellcolor{black!10}- - & 
        \cellcolor{blue!10}$94/100$ & 
        \cellcolor{black!10}- -\\
        \hline
        Drug (M) & \cellcolor{blue!10}$100/100$ & 
        \cellcolor{black!10}- - & 
        \cellcolor{blue!10}$82/100$ & 
        \cellcolor{black!10}- -\\
        \hline
        
    \end{tabular}
    
    \caption{\label{Tab:HstHivOurs2}The statistical outcomes of our \textcolor{cyan}{\texttt{WGAN-GP+G\_EOT+VAE+Buffer}} synthetic dataset.}
\end{table}

\newpage

\end{document}